\documentclass[journal,twoside]{IEEEtran}
\usepackage{amsfonts,bm,booktabs,cite,graphicx,hyperref,multirow,textcomp,url}
\usepackage{algorithm}
\usepackage{subfigure}
\usepackage{algpseudocode}
\usepackage{amsmath}
\usepackage{mathrsfs}
\usepackage{float}
\usepackage{color}
\usepackage{booktabs}
\usepackage{bbding}
\usepackage[flushleft]{threeparttable}

\makeatletter 
\renewcommand{\@thesubfigure}{\hskip\subfiglabelskip}
 \makeatother
\begin{document}
\title{NVC-1B: A Large Neural Video Coding Model}
\author{
	Xihua Sheng$^*$, 
    Chuanbo Tang$^*$,
	Li Li, \IEEEmembership{Member, IEEE},
	Dong Liu, \IEEEmembership{Senior Member, IEEE},
	Feng Wu, \IEEEmembership{Fellow, IEEE}\\
\thanks{
%This paper was received on July 11, 2022; revised on September 23, 2022; accepted on October 29, 2022; 
Date of current version \today.\par 
X. Sheng, C. Tang, L. Li, D. Liu, and F. Wu are with the MOE Key Laboratory of Brain-Inspired Intelligent Perception and Cognition, University of Science and Technology of China, Hefei 230027, China (e-mail: xhsheng@mail.ustc.edu.cn, cbtang@mail.ustc.edu.cn, lil1@ustc.edu.cn, dongeliu@ustc.edu.cn, fengwu@ustc.edu.cn). Corresponding author: Feng Wu.\par $^*$Xihua Sheng and Chuanbo Tang contributed equally to this work.
}
}

\markboth{Submitted to IEEE Transactions on Circuits and Systems for Video Technology}{NVC-1B: A Large Neural Video Coding Model}

\maketitle
\begin{abstract}
The emerging large models have achieved notable progress in the fields of natural language processing and computer vision. However, large models for neural video coding are still unexplored. In this paper, we try to explore how to build a large neural video coding model. Based on a small baseline model, we gradually scale up the model sizes of its different coding parts, including the motion encoder-decoder, motion entropy model, contextual encoder-decoder, contextual entropy model, and temporal context mining module, and analyze the influence of model sizes on video compression performance. Then, we explore to use different architectures, including CNN, mixed CNN-Transformer, and Transformer architectures, to implement the neural video coding model and analyze the influence of model architectures on video compression performance. Based on our exploration results, we design the first neural video coding model with more than 1 billion parameters---NVC-1B. Experimental results show that our proposed large model achieves a significant video compression performance improvement over the small baseline model, and represents the state-of-the-art compression efficiency. We anticipate large models may bring up the video coding technologies to the next level. 
\end{abstract}
\IEEEpeerreviewmaketitle

\section{Introduction}
The popularity of video applications, such as short video-sharing platforms, video conferences, and streaming television, has made the amount of video data increase rapidly. The large data amount brings large costs for video transmission and storage. Therefore, it is urgent to compress videos efficiently to reduce video data amount.\par
To decrease the costs of video transmission and storage, various advanced coding technologies have been proposed, including intra/inter-frame prediction, transform, quantization, entropy coding, and loop filters. These coding technologies have led to the development of a series of video coding standards over the past decades, such as H.264/AVC~\cite{wiegand2003overview}, H.265/HEVC~\cite{sullivan2012overview}, and H.266/VVC~\cite{bross2021overview}. These video coding standards have significantly improved video compression performance. \par

Although traditional video coding standards have achieved great success, their compression performance is increasing at a slower speed than that of video data amount. To break through the bottleneck of video compression performance growth, researchers have begun to explore end-to-end neural video coding in recent years. Existing end-to-end neural video coding (NVC) models can be roughly classified into four classes: volume coding-based~\cite{Habibian_2019_ICCV,sun2020high}, conditional entropy modeling-based~\cite{liu2020conditional,DBLP:conf/nips/MentzerTMCHLA22}, 
implicit neural representation-based~\cite{chen2021nerv,chen2023hnerv,zhao2023dnerv,kwan2024hinerv}, motion compensated prediction-based~\cite{liu2020learned,Rippel_2021_ICCV,lu2020end,lin2020m,hu2021fvc,agustsson2020scale,cheng2019learning,rippel2019learned,djelouah2019neural,yang2021learning,liu2021deep,liu2020neural,yilmaz2021end,chen2019learning,lin2022dmvc,guo2023learning,guo2023enhanced,yang2022advancing,sheng2022temporal,wang2023butterfly,sheng2024spatial,sheng2024vnvc,li2021deep,li2022hybrid,ho2022canf,jin2023learned,lin2023multiple,li2023neural}. 
With the development of neural networks in the fields of natural language processing and computer vision, recent neural video coding models~\cite{sheng2024spatial,li2023neural,li2024neural} have surpassed the reference software of H.266/VVC standard under certain coding conditions. However, the development speed of neural video coding is lower than that of natural language processing and computer vision. One typical example is that the emerging large models for natural language processing or computer vision have achieved great success, but large models for neural video coding are still unexplored.\par

Large models for natural language processing and computer vision have attracted much attention recently. Large language models (LLMs) and large vision models (LVMs) are commonly first built on small models and then their model sizes are gradually scaled up. Following the scaling law~\cite{kaplan2020scaling}, increasing the model sizes can bring impressive task performance gains. In terms of LLMs, their model sizes can reach hundreds to thousands of billion (B). For example, the representative LLMs---GPT-3~\cite{brown2020language} and GPT-4~\cite{achiam2023gpt} developed by OpenAI contain 175B parameters and 1800B parameters, respectively. Their natural language understanding ability and human language generation ability are greatly improved. In terms of LVMs, their model sizes are much smaller than those of LLMs. For example, the Segment Anything Model (SAM)~\cite{kirillov2023segment} used for image segmentation and the Depth Anything Model~\cite{yang2024depth} only contain 636 million (M) parameters and 335M parameters. Even the largest Vision Transformer (ViT)~\cite{dosovitskiy2020image} models have only recently grown from a few hundred million to 22B~\cite{dehghani2023scaling}. Although the model sizes of current LVMs are much smaller than those of LLMs, larger model sizes for vision models still greatly improve computer vision task performance~\cite{liu2022swin}.\par

Witnessing the success of LLMs and LVMs, in this paper, we try to explore the effectiveness of large model for neural video coding. We regard a small model--DCVC-SDD~\cite{sheng2024spatial} as our baseline, which is based on the typical conditional coding architecture~\cite{li2021deep, sheng2022temporal, li2022hybrid, li2023neural} and has a higher compression performance than the reference software of H.266/VVC---VTM-13.0~\cite{VTM} under certain testing conditions. We first analyze the influence of model sizes on compression performance. Specifically, we gradually scale up the model sizes of its motion encoder-decoder, motion entropy model, contextual encoder-decoder, contextual entropy model, and temporal context mining module. Then, we analyze the influence of model architectures on compression performance. Specifically, we try to replace the CNN architecture with mixed CNN-Transformer or  Transformer-only architecture.  Finally, we build a large neural video coding model with 1B parameters---NVC-1B.\par
Our contributions are summarized as follows:
\begin{itemize}
    \item We scale up the model sizes of different parts of a neural video coding model and analyze the influence of model sizes on video compression performance.

    \item We use different architectures to implement the neural video coding model and analyze the influence of model architectures on video compression performance.

    \item Based on our exploration results, we propose the first large neural video coding model with more than 1B parameters---NVC-1B and achieve a significant video compression performance gain over the small baseline model.

\end{itemize}
The remainder of this paper is organized as follows.  Section~\ref{sec:related_work} gives a review of related work about LLM, LVM, and neural video coding. Section~\ref{sec:overview} introduces the framework overview of our proposed NVC-1B. Section~\ref{sec:methodology} analyzes the influence of model sizes and model architectures on compression performance in detail. Section~\ref{sec:experiments} reports the compression performance of NVC-1B and gives some analyses. Section~\ref{sec:conclusion} gives a conclusion of this paper.

\section{Related Work}\label{sec:related_work}
\subsection{Large Language Model}
A large language model (LLM) refers to a language model containing billions of parameters with the goal of natural language
understanding and human language generation.  Recently, a variety of LLMs have emerged, including those developed by OpenAI (GPT-3~\cite{brown2020language} and GPT-4~\cite{achiam2023gpt}), Google (GLaM~\cite{du2022glam}, PaLM~\cite{chowdhery2023palm}, PaLM-2~\cite{anil2023palm}, and Gemini~\cite{team2023gemini}), and Meta (LLaMA-1~\cite{touvron2023llama}, LLaMA-2~\cite{touvron2023llama}, and LLaMA-3~\cite{meta2024introducing}). These models are commonly first built upon the small language models. Then, under the guidance of the scaling law~\cite{kaplan2020scaling}, they largely scale the model sizes. For example, GPT-3 contains 175B parameters and GPT-4 even contains 1800B parameters.
Owing to the massive scale in model sizes, LLMs 
have achieved significant performance across various tasks. They can better understand natural languages and generate high-quality texts based on
interactive contexts, such as prompts.

\subsection{Large Vision Model}
The success of LLMs leads to the rise of large vision models (LVMs). Similar to LLMs, scaling up the model sizes of small vision models has also shown improved performance for various tasks. Google researchers~\cite{dehghani2023scaling} proposed Vision Transformer (ViT)~\cite{dosovitskiy2020image} for image and video modeling. Its model sizes gradually increase from several hundred million to 4B~\cite{chen2022pali}. Recently, its model size has even been extended to 22B~\cite{dehghani2023scaling} and achieved the current
state-of-the-art ranging from (few-shot) classification to
dense output vision tasks. Based on the ViT backbone, LAION researchers~\cite{cherti2023reproducible} have investigated scaling laws for contrastive language-image pre-training (CLIP)~\cite{radford2021learning}. The largest model based on ViT-G/14 can reach 1.8B parameters. Their investigation shows that scaling up the model size of CLIP can improve the performance of zero-shot classification of downstream vision tasks. Also based on ViT, Meta researchers proposed a Segment Anything Model (SAM)~\cite{kirillov2023segment} for image segmentation. The largest SAM model has about 636M parameters and has achieved excellent segmentation accuracy on many benchmark datasets. Tiktok researchers proposed a Depth Anything Model~\cite{yang2024depth} with 335M parameters for monocular image depth estimation and has shown impressive depth estimation accuracy.  Although the model sizes of current LVMs are much smaller than those of LLMs, these pioneering works of LVMs still verify the benefits of scaling up the model sizes for visual tasks. However, no research has explored the effectiveness of scaling up the model size of neural video coding models.

\subsection{Neural Video Coding}
Neural video coding~\cite{jin2023learned,Habibian_2019_ICCV,wu2018video,rippel2019learned,DBLP:conf/cvpr/LuO0ZCG19,chen2019learning,lu2020end,liu2020learned,liu2020neural,liu2020conditional,agustsson2020scale,lin2020m,hu2021fvc,yilmaz2021end,Rippel_2021_ICCV,chen2021nerv,li2021deep,yang2021learning,hu2022coarse,liu2022end,shi2022alphavc,DBLP:conf/nips/MentzerTMCHLA22,xiang2022mimt,lin2022dmvc,yang2022advancing,guo2023learning,xu2023bit,guo2023enhanced,chen2023hnerv,kim2023c3,sheng2022temporal,li2024neural,tang2024offline,lu2024deep,sheng2024vnvc,du2024cgvc,kwan2024hinerv} has explored a new direction for video compression in recent years. Among different kinds of coding schemes, motion compensation-based schemes have achieved state-of-the-art compression performance. \par

DVC~\cite{DBLP:conf/cvpr/LuO0ZCG19} is a pioneering scheme of motion compensation-based neural video coding. 
It follows the traditional hybrid video coding framework but implements main modules with neural networks, such as motion estimation, motion compression, motion compensation, residual compression, and entropy models.
Based on DVC, subsequent schemes mainly focus on how to increase the accuracy of temporal prediction. For example, 
Lin et al.~\cite{lin2020m} proposed M-LVC that introduces multiple reference frames for motion compensation and motion vector prediction. Agustsson et al.~\cite{agustsson2020scale} proposed SSF that generalizes typical motion vectors to a scale-space flow for better handling complex motion. Hu et al.~\cite{hu2021fvc} proposed FVC that shifts temporal prediction from pixel domain to feature domain using deformable convolution~\cite{dai2017deformable}.\par
%-------------------------------------------------------------------------
\begin{figure*}[t]
  \centering
   \includegraphics[width=0.7\linewidth]{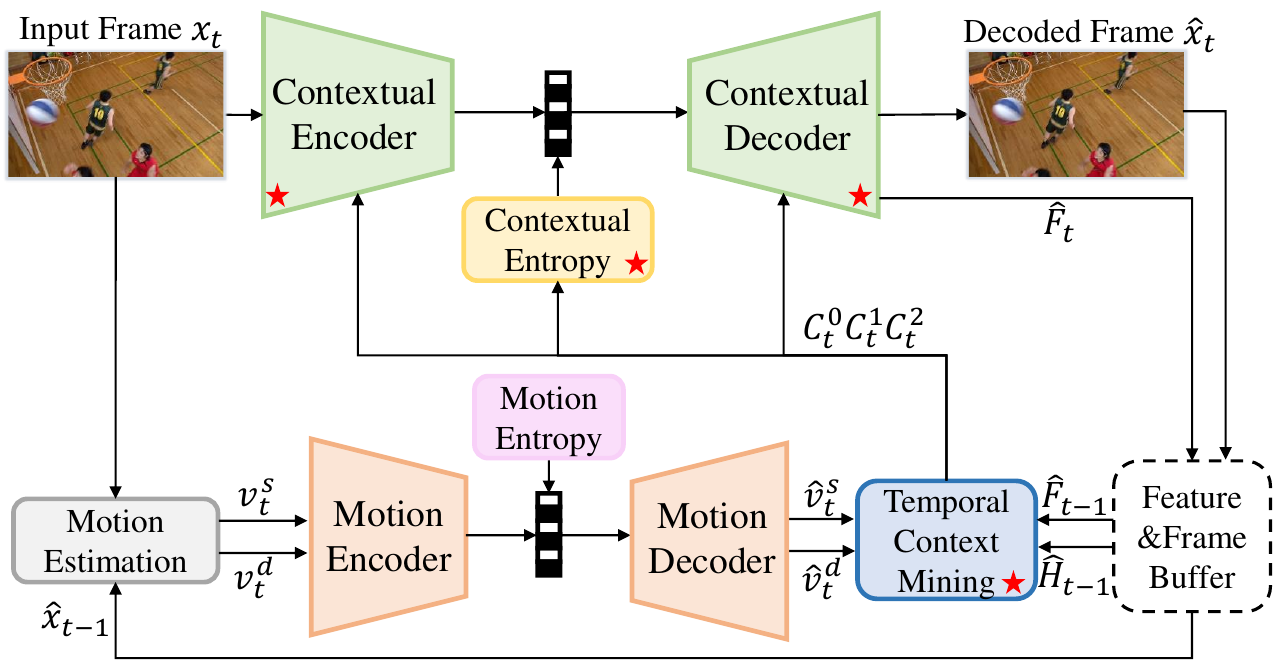}
      \caption{Overview of our proposed large neural video coding model--NVC-1B. We explore to scale up the model size motion encoder-decoder, motion entropy model, contextual encoder-decoder, contextual entropy model, and temporal context mining module. Based on our exploration results, we allocate most model parameters to the modules marked with red stars.}
   \label{fig:framework}
\end{figure*}
%-------------------------------------------------------------------------
Different from DVC-based schemes, DCVC~\cite{li2021deep} shifts the typical residual coding paradigm to a conditional coding paradigm. Regarding temporal prediction as a condition, DCVC feeds the condition into a contextual encoder-decoder, allowing the networks to learn how to reduce temporal redundancy automatically rather than performing explicit subtraction operations. Based on DCVC, Sheng et al.~\cite{sheng2022temporal} proposed DCVC-TCM that designs a temporal context mining module to generate multi-scale temporal contexts.  This work greatly improved the compression performance of neural video coding and began to make fair comparisons with standard reference software of traditional codecs. Following DCVC-TCM, DCVC-HEM~\cite{li2022hybrid}, DCVC-DC~\cite{li2023neural}, and DCVC-SDD~\cite{sheng2024spatial} were further proposed. They focus on introducing more temporal conditions to utilize temporal correlation, such as latent-prior, decoding parallel-friendly spatial-prior, and long-term temporal prior. Nowadays, the compression performance of the DCVC series has exceeded that of the reference software of H.266/VVC~\cite{bross2021developments}.\par
Existing neural video coding models commonly have small model sizes. In this work, we try to scale up the model size of a neural video coding model to explore the influence of model size on video compression performance.

\section{Overview}\label{sec:overview}
We build the large neural video coding model--NVC-1B based on our small baseline model DCVC-SDD~\cite{sheng2024spatial}. 
We first give a brief overview of the framework of our NVC-1B. 

\subsubsection{Motion Estimation}
Similar to our small baseline model DCVC-SDD, we use the structure and detail decomposition-based motion estimation method to estimate the motion vectors between adjacent video frames~\cite{sheng2024spatial}. Specifically, we first decompose the current frame $x_t$ and the reference frame $\hat{x}_{t-1}$ into structure components ($x_{t}^s$, $\hat{x}_{t-1}^s$) and detail components  ($x_{t}^d$, $\hat{x}_{t-1}^d$) . Then, we estimate the motion vectors ($v_{t}^s$, $v_{t}^d$)  of the structure and detail components respectively using a pre-trained SpyNet~\cite{ranjan2017optical}.

\subsubsection{Motion Encoder-Decoder}
We use an autoencoder-based motion encoder-decoder to compress and reconstruct the estimated motion vectors (MVs) jointly, as shown in Fig.~\ref{fig:framework}. Specifically, we first perform channel-wise concatenation to the input motion vectors $v_{t}^s$ and $v_{t}^d$. Then, the MV encoder compresses them into a compact latent representation $m_t$ with the size of  $H/16 \times W/16 \times C_m$. $H$ is the height and $W$ is the width of the input motion maps.  $C_m$ is the number of channels of the compact latent representation $m_t$. After quantization, the quantized latent representation $\hat{m}_t$ is signaled into a bitstream using the arithmetic encoder. After receiving the transmitted bitstream, the arithmetic decoder reconstructs the bitstream back to the quantized latent representation $\hat{m}_t$. The MV decoder then decompresses $\hat{m}_t$ back to the reconstructed motion vectors  $\hat{v}_{t}^s$ and $\hat{v}_{t}^d$. 

\subsubsection{Temporal Context Mining}
The temporal context mining module is essential for conditional-based neural video coding schemes~\cite{sheng2024spatial,sheng2024vnvc,sheng2022temporal,jin2023learned,li2021deep,li2022hybrid,li2023neural,li2024neural}. Given a reference feature $\hat{F}_t$, the temporal context mining module first uses a feature pyramid to extract multi-scale features from $\hat{F}_t$. Then, it uses muli-scale motion vectors to perform feature-based motion compensation to these features and learn multi-scale temporal contexts $\Tilde{C}_t^0$,$\Tilde{C}_t^1$,$\Tilde{C}_t^2$. To handle motion occlusion, following our baseline~\cite{sheng2024spatial}, we use a ConvLSTM-based long-term reference generator to accumulate the historical information of each reference feature and learn a long-term reference feature $\hat{H}_{t-1}$. Then, we fuse $\hat{H}_{t-1}$ with  $\Tilde{C}_t^0$,$\Tilde{C}_t^1$,$\Tilde{C}_t^2$ to generate long short-term fused temporal contexts $C_t^0$,$C_t^1$,$C_t^2$.

\subsubsection{Contextual Encoder-Decoder}
We use an autoencoder-based contextual encoder and decoder to compress and reconstruct the input frame $x_t$. The contextual encoder compresses $x_t$ into a compact latent representation $y_t$ with the size of $H/16 \times W/16 \times C_y$. $C_y$ is the number of channels of the compact latent representation $y_t$. In the encoding procedure, multi-scale temporal contexts $C_t^0$,$C_t^1$,$C_t^2$ learned by the temporal context mining module are channel-wise concatenated into the contextual encoder to reduce temporal redundancy. Then, quantization is performed to $y_t$ and the arithmetic encoder converts the quantized latent representation $\hat{y}_t$ to a bitstream. After receiving the transmitted bitstream, the arithmetic decoder reconstructs it to $\hat{y}_t$. The contextual decoder decompresses $\hat{y}_t$ to a reconstructed frame $\hat{x}_t$. In the decoding procedure, the multi-scale temporal contexts $C_t^0$,$C_t^1$,$C_t^2$ are also concatenated into the contextual decoder. Before obtaining $\hat{x}_t$, an intermediate feature $\hat{F}_t$ of the contextual decoder with the size of $H \times W \times C_F$ is regarded as the reference feature for encoding/decoding the next frame. 

\subsubsection{Entropy Model}
We use the factorized entropy model~\cite{DBLP:conf/iclr/BalleLS17} for hyperprior and the Laplace distribution~\cite{DBLP:conf/iclr/BalleMSHJ18} to model the motion and contextual compact latent representations $\hat{m}_t$ and $\hat{y}_t$. When estimating the mean and scale of the Laplace distribution of $\hat{m}_t$ and $\hat{y}_t$, we combine the hyperprior, latent prior, and the spatial prior generated by the quadtree partition-based spatial entropy model~\cite{sheng2024spatial,li2023neural} together. For $\hat{y}_t$, we also introduce a temporal prior learned from the smallest-resolution temporal context $C_{t}^{2}$. 

\section{Methodology}\label{sec:methodology}
To explore the influence of model size on compression performance, we gradually scale up the model sizes of different parts of our small neural video coding model~\cite{sheng2024spatial}, including its motion encoder-decoder, motion entropy model, contextual encoder-decoder, contextual entropy model, and temporal context mining module.  Since most existing neural video coding models use a pre-trained optical flow model for motion estimation and focus on designing other coding parts, we do not scale up the model size of the motion estimation module in this work. To explore the influence of model architecture on compression performance, we use different architectures such as mixed CNN-Transformer and Transformer architectures to implement the neural video coding model. In the exploration procedure, we regard our small baseline model---DCVC-SDD (21M parameters) without multiple-frame cascaded finetune as the anchor to reduce training time.
%-------------------------------------------------------------------------
\begin{figure}[t]
  \centering
   \includegraphics[width=\linewidth]{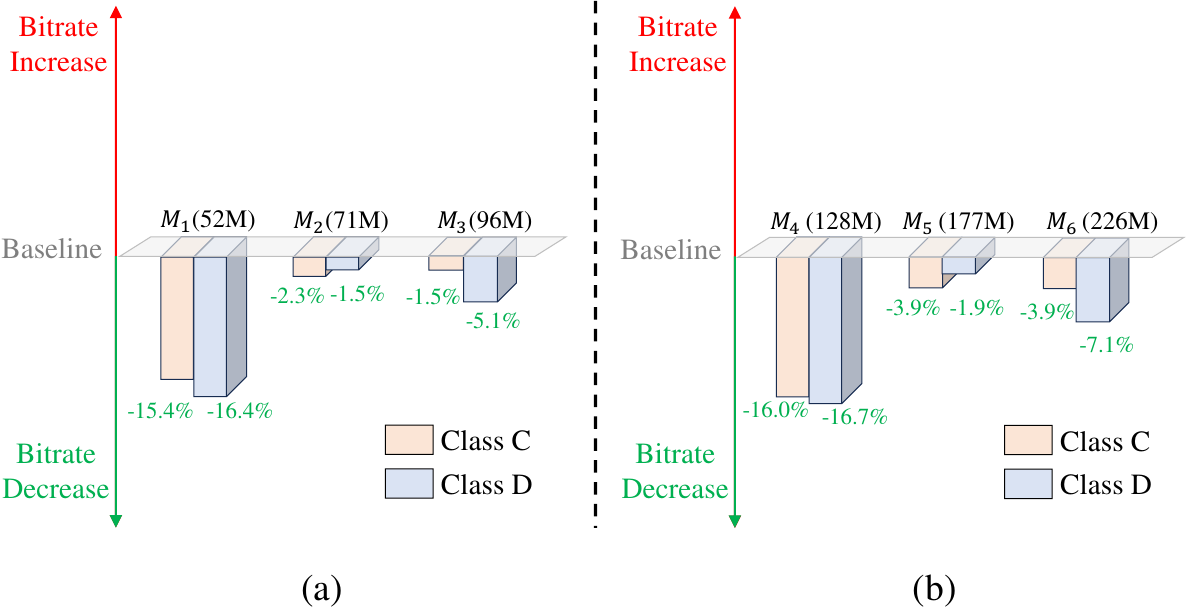}
      \caption{(a) Effectiveness of scaling up the model size of motion encoder-decoder. Based on our small baseline model~\cite{sheng2024spatial} with 21M parameters, three models ($M_1$, $M_2$, $M_3$) with 52M, 71M, and 96M parameters respectively are built. (b) Effectiveness of scaling up the model size of motion entropy model. Based on $M_1$, $M_2$, $M_3$, three models ($M_4$, $M_5$, $M_6$) with 128M, 177M, and 226M parameters respectively are built. When calculating the BD-rate, the anchor is our small baseline model.}
   \label{fig:mv_encoder_decoder_result}
\end{figure}

\subsection{Influence of Model Size}\label{sec:model_size}
\subsubsection{Scaling Up for Motion Encoder-Decoder}\label{sec:mv_encoder_decoder}
We increase the number of intermediate feature channels and insert more residual blocks to scale up the model size of the motion encoder and decoder. We build three models ($M_1$, $M_2$, $M_3$) with 52M, 71M, and 96M parameters, respectively. The comparison results as illustrated in Fig.~\ref{fig:mv_encoder_decoder_result}(a) indicate that a slight increase in the model size of the motion encoder and decoder results in performance gains. However, continuously increasing the model size of the motion encoder-decoder will reduce the performance gain. For example, when the model size of the motion encoder-decoder is increased to 52M,  15.4\% performance gain can be achieved for the HEVC Class C dataset. However, if we further increase its model size to 96M parameters, the performance gain drops to 1.5\%. The results indicate that the motion encoder-decoder is not always the larger the better. More analysis can be found in Section~\ref{ablation3}. In addition, we observe that although larger motion encoder-decoders bring performance gain, the training processes become unstable and training crashes occur more frequently.
%-------------------------------------------------------------------------
\begin{figure}[t]
  \centering
   \includegraphics[width=\linewidth]{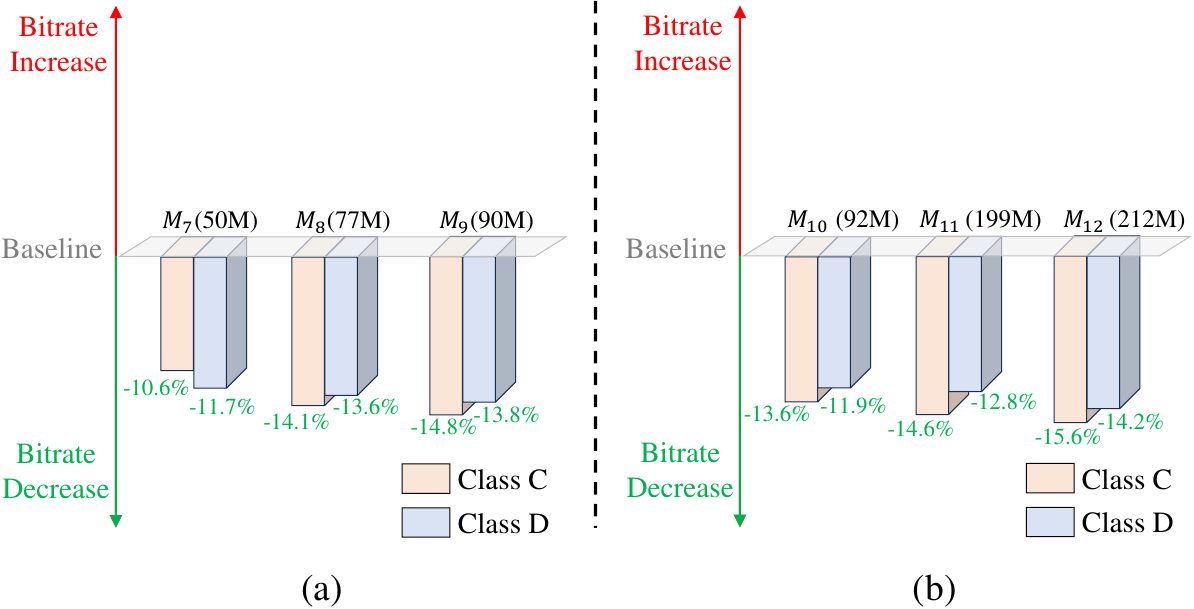}
      \caption{(a) Effectiveness of scaling up the model size of contextual encoder-decoder.  Based on our small baseline model with 21M parameters, three models ($M_7$, $M_8$, $M_9$) with 50M, 77M, and 90M parameters respectively are built. (b) Effectiveness of scaling up the model size of contextual entropy model. Based on $M_7$, $M_8$, $M_9$, three models ($M_{10}$, $M_{11}$, $M_{12}$) with 92M, 199M, and 212M parameters respectively are built. When calculating the BD-rate, the anchor is our small baseline model.}
   \label{fig:contextual_encoder_decoder_result}
\end{figure}
%-------------------------------------------------------------------------
\subsubsection{Scaling Up for Motion Entropy Model}Based on the abovementioned $M_1$, $M_2$, $M_3$ models, we first increase the channel number of the motion latent representation $\hat{m}_t$ and associated hyperprior $\hat{z}_t^m$. Then, we scale up the model size of motion hyper-encoder, hyper-decoder, and quadtree partition-based spatial context models~\cite{sheng2024spatial, li2023neural} by increasing the number of intermediate feature channels. We build three models ($M_4$, $M_5$, $M_6$) with 128M, 177M, and 226M parameters, respectively. The comparison results as illustrated in Fig.~\ref{fig:mv_encoder_decoder_result}(b) indicate that the increase in the model size of the motion entropy model can bring a little compression performance gain. For example, based on the $M_1$ model with 52M parameters, the $M_4$ model improves the performance gain from 15.4\% to 16.0\% for the HEVC Class C dataset. However, although $M_4$, $M_5$, $M_6$ models can further improve performance,  similar to $M_1$, $M_2$, $M_3$ models, training crashes occur more frequently.

\subsubsection{Scaling Up for Contextual Encoder-Decoder}
Based on our small baseline model with 21M parameters, we increase the number of intermediate feature channels and insert more residual blocks to scale up the model size of the contextual encoder and decoder. We build three models ($M_7$, $M_8$, $M_9$) with 50M, 77M, and 90M parameters, respectively. As presented in Fig.~\ref{fig:contextual_encoder_decoder_result}(a), increasing the model size of the context encoder and decoder can effectively improve compression performance.  Different from the motion encoder and decoder, further increasing the model size of the contextual encoder and decoder can result in sustained and stable performance gains. For example, when the model size of the contextual encoder and decoder is increased to 50M, 10.6\% performance gain can be achieved for the HEVC Class C dataset. When the model size of the contextual encoder and decoder is further increased to 90M, the performance gain can reach 14.8\% for the HEVC Class C dataset. The results show that a larger contextual encoder-decoder can improve the transform capability. More analysis can be found in Section~\ref{ablation1}.
%-------------------------------------------------------------------------
\begin{figure}[t]
  \centering
   \includegraphics[width=\linewidth]{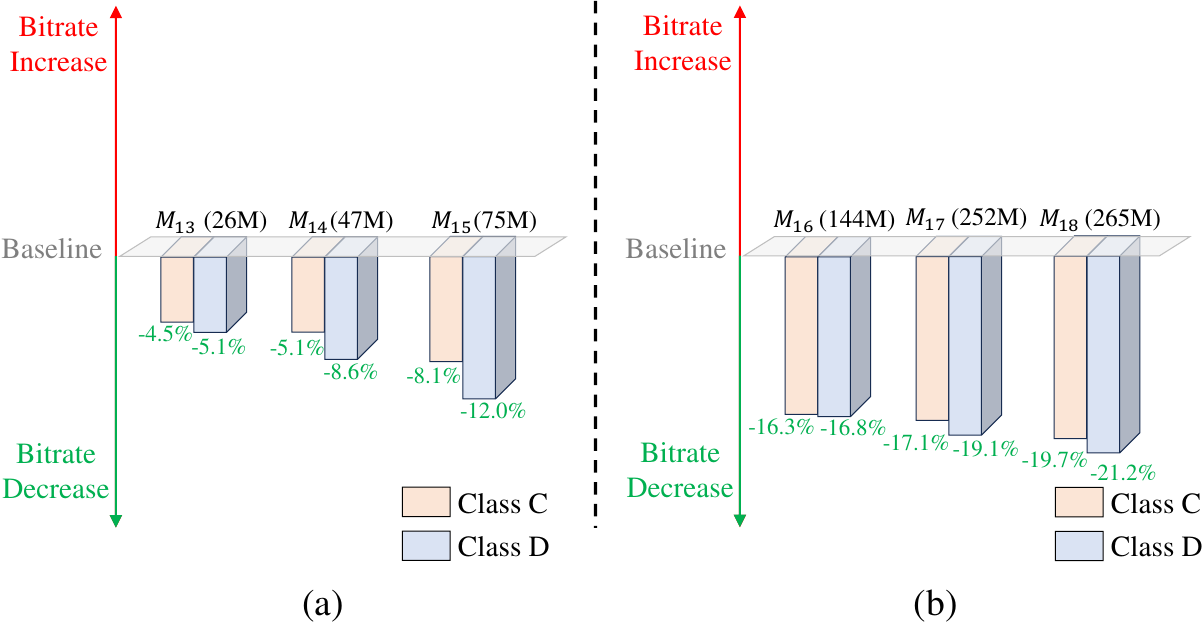}
      \caption{(a) Effectiveness of scaling up the model size of temporal context mining module.  Based on our small baseline model with 21M parameters, three models ($M_{13}$, $M_{14}$, $M_{15}$) with 26M, 47M, and 75M parameters respectively are built. (b) To explore whether the gain of scaling up the temporal context mining module can be superimposed with the gain of scaling up the contextual encoder-decoder and contextual entropy model, based on $M_{10}$, $M_{11}$, $M_{12}$, three models ($M_{16}$, $M_{17}$, $M_{18}$) with 144M, 252M, and 265M parameters respectively are built. When calculating the BD-rate, the anchor is our small baseline model.}
   \label{fig:TCM_results}
\end{figure}
%-------------------------------------------------------------------------
\subsubsection{Scaling Up for Contextual Entropy Model}
Based on $M_7$, $M_8$, $M_9$ models, we continue to scale up the model size of their contextual entropy models. We increase the channel number of the contextual latent representation $\hat{y}_t$ and associated hyperprior $\hat{z}_t^y$. In addition, we increase the channel number of intermediate features of the contextual entropy model, including the contextual hyper-encoder, hyper-decoder, and quadtree partition-based spatial context models~\cite{sheng2024spatial, li2023neural}. We build three models ($M_{10}$, $M_{11}$, $M_{12}$) with 92M, 199M, and 212M parameters, respectively. As presented in Fig.~\ref{fig:contextual_encoder_decoder_result}(b), increasing the model size of the context encoder and decoder can bring stable compression performance improvement. For example, when the model size of $M_9$ model is increased from 90M to 212M by scaling up its contextual entropy model, an additional 0.8\% performance gain can be achieved for the HEVC Class C dataset. The results show that a larger contextual entropy model can improve the entropy modeling of the contextual latent representation. 

\subsubsection{Scaling Up for Temporal Context Mining}
Temporal context mining is an intermediate link between the motion encoder-decoder and contextual encoder-decoder. To explore the influence of its model size, we increase the number of intermediate feature channels $N$ and insert more residual blocks to scale up its model size. Based on our small baseline model with 21M parameters, we build three models ($M_{13}$, $M_{14}$, $M_{15}$) with 26M, 47M, and 75M parameters, respectively. As shown in Fig.~\ref{fig:TCM_results}(a), scaling up the model size of the temporal context mining module can improve compression performance stably. For example, when the model size is increased from 26M to 75M, the performance gain can be increased from 5.1\% to 12.0\% for the HEVC Class D dataset. To explore whether the gain of scaling up the temporal context mining module can be superimposed with the gain of scaling up the contextual encoder-decoder and contextual entropy model, based on $M_{10}$, $M_{11}$, $M_{12}$, we build another three models ($M_{16}$, $M_{17}$, $M_{18}$) with 144M, 252M, and 265M parameters, respectively. As shown in Fig.~\ref{fig:TCM_results}(b), based on the model with a larger contextual encoder-decoder and contextual entropy model, scaling up the model size of the temporal context mining module can bring additional performance gain. For example, based on $M_{12}$ with 212M parameters, increasing its temporal context mining module to build the model $M_{18}$ with 265M parameters can make the performance gain increase from 14.2\% to 21.2\% for the HEVC Class D dataset. The results show that a larger temporal context mining module can help make full use of the temporal correlation. More analysis can be found in Section~\ref{ablation2}.

\subsection{Influence of Model Architecture}\label{sec:model_arc}
In addition to exploring the influence of model size, we also explore the influence of model architecture. In Section~\ref{sec:model_size}, all the models adopt CNN architectures. In this section, we try to replace the CNN architecture with mixed CNN-Transformer or Transformer architecture. Since the experimental results presented in Section~\ref{sec:model_size} indicate that scaling up the model sizes of contextual encoder-decoder, contextual entropy model, and temporal context mining module can bring stable compression performance improvement, we try new model architectures on these modules that have been proven to work. 
%-------------------------------------------------------------------------
\begin{figure}[t]
  \centering
   \includegraphics[width=\linewidth]{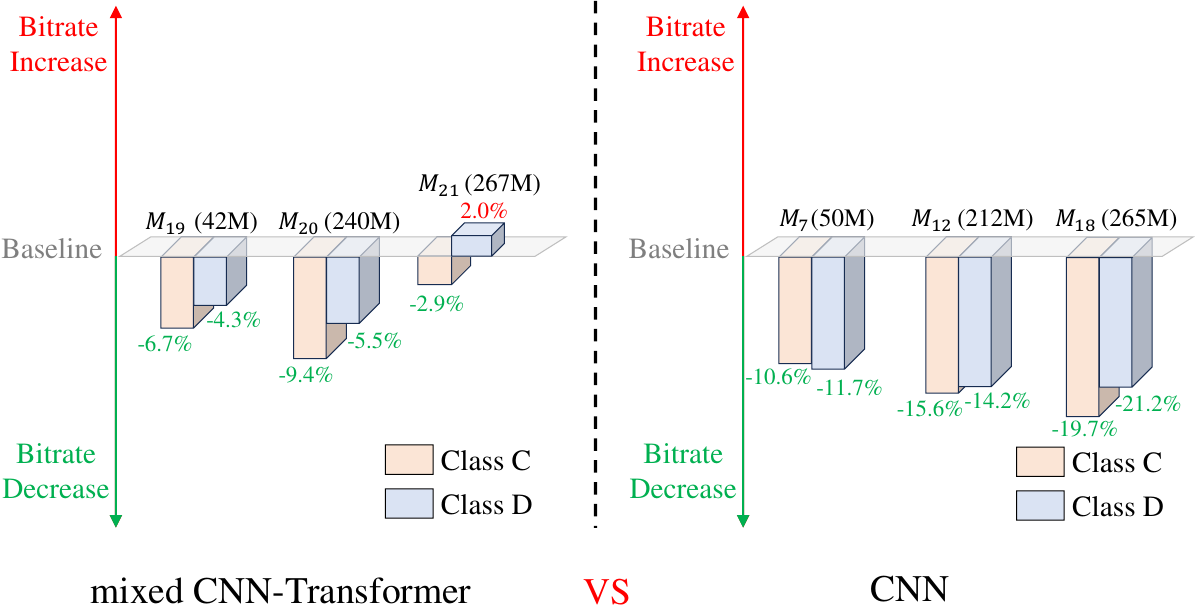}
      \caption{Effectiveness of the mixed CNN-Transformer architecture.  Based on our small baseline model with 21M parameters, we gradually insert SwinTransformer layers~\cite{liu2021Swin} into the contextual encoder-decoder, contextual entropy model, and temporal context mining module with CNN architectures and build three models ($M_{19}$, $M_{20}$, $M_{21}$) with 42M, 240M, and 267M parameters respectively.   
      For comparison, we also list the compression results of their CNN-architecture counterparts with similar model sizes. 
      When calculating the BD-rate, the anchor is our small baseline model.}
   \label{fig:CNN-Transformer_results}
\end{figure}
%-------------------------------------------------------------------------
\subsubsection{Mixed CNN-Transformer Architecture}
In terms of the mixed CNN-Transformer architecture, we build three models $M_{19}$, $M_{20}$, $M_{21}$. For model $M_{19}$, we replace the residual blocks between the convolutional layers of the contextual encoder and decoder with SwinTransformer layers~\cite{liu2021Swin} to build a contextual encoder-decoder with mixed CNN-Transformer architecture. We scale up its model size to 42M to compare it with its CNN-architecture counterpart---$M_{7}$ model (50M). For model $M_{20}$, based on $M_{19}$, we continue to insert SwinTransformer layers into the CNN-based entropy model to build a contextual entropy model with mixed CNN-Transformer architecture.  We scale up its model size to 240M to compare it with its CNN architecture counterpart---$M_{12}$ model (212M). For model $M_{21}$, based on $M_{20}$, we insert a SwinTransformer layer after each residual block of the temporal context mining module. We scale up its model size to 267M to compare it with its CNN architecture counterpart---$M_{18}$ model (265M). As illustrated in Fig.~\ref{fig:CNN-Transformer_results}(a),  for model $M_{19}$ and model $M_{20}$, inserting the Transformer layers into the original CNN architecture can bring performance gain over the small baseline model for its global feature extraction ability. This phenomenon is consistent with previous work~\cite{lu2021transformer,qian2022entroformer,liu2023learned} on image coding based on Transformer.
However, when compared with their CNN-architecture counterparts of similar model sizes, CNN-architecture counterparts can obtain higher video coding performance gain. For example, comparing $M_{20}$ and $M_{12}$, which both have larger contextual encoder-decoder and larger contextual entropy model, the performance gain of $M_{12}$ is 15.6\% but that of $M_{20}$ is only 9.4\%  for the HEVC Class C dataset. For model $M_{21}$, inserting SwinTransformer layers into the temporal context mining module brings performance loss. The compression performance gain drops from 9.4\% achieved by $M_{20}$ to 2.9\%. The results indicate that the global feature extraction ability of Transformer layers may not be suitable for learning multi-scale temporal contexts.
 
 %-------------------------------------------------------------------------
\begin{figure}[t]
  \centering
   \includegraphics[width=\linewidth]{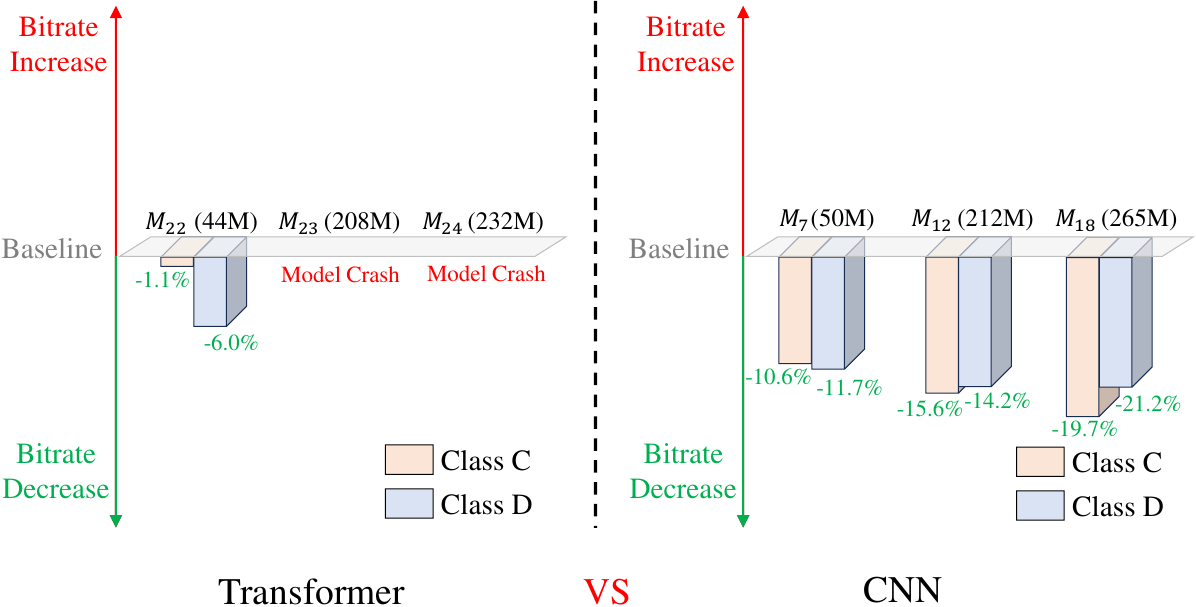}
      \caption{Effectiveness of the Transformer architecture.  Based on our small baseline model with 21M parameters, we gradually replace the convolutional layers, residual blocks, depth blocks of contextual encoder-decoder, contextual entropy model, and temporal context mining module with SwinTransformers layer~\cite{liu2021Swin} and build three models ($M_{19}$, $M_{20}$, $M_{21}$) with 44M, 208M, and 232M parameters respectively. For comparison, we also list the compression results of their CNN-architecture counterparts with similar model sizes.
      When calculating the BD-rate, the anchor is our small baseline model.}
   \label{fig:Transformer_results}
\end{figure}
%-------------------------------------------------------------------------
\subsubsection{Transformer Architecture}
In terms of the Transformer architecture, we build three models $M_{22}$, $M_{23}$, $M_{24}$. For model $M_{22}$, we replace all the convolutional layers and residual blocks of the contextual encoder and decoder with SwinTransformer layers~\cite{liu2021Swin} to build a contextual encoder-decoder with Transformer architecture. We scale up its model size to 44M to compare it with its CNN-architecture counterpart---$M_{7}$ model (50M). For model $M_{23}$, based on $M_{22}$, we further replace the convolutional layers and depth blocks in the contextual entropy model with SwinTransformer layers to build a contextual entropy model with Transformer architecture.  We scale up its model size to 208M to compare it with its CNN architecture counterpart---$M_{12}$ model (212M). For model $M_{24}$, based on $M_{23}$, we replace all the convolutional layers and residual blocks of the temporal context mining module with SwinTransformer layers. We scale up its model size to 232M to compare it with its CNN architecture counterpart---$M_{18}$ model (265M). As illustrated in Fig.~\ref{fig:Transformer_results}, for model $M_{22}$, the Transformer-based contextual encoder-decoder can bring a little performance gain over the small baseline model. However, when compared with its CNN architecture counterpart $M_7$ with similar model sizes, its performance gain is much smaller. For model $M_{23}$ and model $M_{24}$, the Transform-only-based contextual entropy model and temporal context mining module make the model crash. Compared with the small baseline model, there is a PSNR loss of more than 4dB, which makes it difficult to calculate the BD-rate values. The results indicate that the convolutional layer is necessary for contextual entropy model and temporal context mining module. 

\begin{table}[t]
\caption{Number of parameters for each coding module of our proposed NVC-1B.} 
  \centering
\scalebox{1}{
\begin{threeparttable}
\begin{tabular}{l|c}
\toprule[1.5pt]
               & Number of Parameters\\ \hline
Motion Estimation            &0.96M\\ \hline
Motion Encoder-Decoder       &0.81M\\ \hline
Motion Entropy Model         &2.50M\\ \hline
Contextual Encoder-Decoder   &504.59M\\ \hline
Contextual Entropy Model     &435.88M\\ \hline
Temporal Context Mining      &411.21M\\ \hline 
Totally                      &1355.95M (1.36B)\\
\bottomrule[1.5pt]
\end{tabular}
  % \begin{tablenotes}
  %  \item \footnotesize \dag DCVC-SDD~\cite{sheng2024spatial} is our small baseline model.

  % \end{tablenotes}
\end{threeparttable}}
\label{table:paramters_all}
\end{table}

\subsection{Summary of Exploration Results}
Based on the abovementioned exploration results, we make the following summary:
\begin{itemize}
    \item Slightly scaling up the model size of the motion encoder-decoder and motion entropy model brings compression performance gains, while further increasing the model size can lead to performance degradation.

    \item Scaling up the model sizes of contextual encoder-decoder, contextual entropy model, and temporal context mining module can bring continuous compression performance improvement. 
    \item With a similar model size, the video coding model with CNN architecture can achieve higher compression performance than that with mixed CNN-Transformer and Transformer architectures.

\end{itemize}
According to the summary of exploration results, we build a large neural video coding model---NVC-1B with CNN architecture.  Under the limited GPU memory condition, and in order to ensure the stability of training, we allocate most of the model parameters to the contextual encoder-decoder, contextual entropy model, and temporal context mining module. The number of parameters for each coding module of our proposed NVC-1B is listed in Table~\ref{table:paramters_all}.

\subsection{Model Training}
We design an elaborate training strategy for our proposed NVC-1B model, ensuring that each of its modules is fully trained. 
Table.~\ref{table:training_stategy_rgb} lists the detailed training stages. The training stages can be classified into 6 classes according to the training loss functions: $L_{t}^{meD}$, $L_{t}^{meRD}$, $L_{t}^{recD}$, $L_{t}^{recRD}$, $L_{t}^{all}$, and $ L_{T}^{all}$. Among them, when using $L_{t}^{meD}$ and $L_{t}^{meRD}$ as the loss functions, we only train the motion parts (Inter). When using $L_{t}^{recD}$ and $L_{t}^{recRD}$ as the loss functions, we only train the temporal context mining and contextual parts (Rec). When using $L_{t}^{all}$ or $L_{T}^{all}$ as the loss function, we train all parts of the model (All).
%-------------------------------------------------------------------------
\begin{table}[t]
\caption{Training strategy of our model for encoding RGB videos when the distortion is measured by RGB PSNR.}
\centering
\begin{tabular}{c|c|c|c|c}
\toprule[1.5pt]
Frames  & Network   & Loss                          &Learning Rate        &Epoch \\ \hline
2     & Inter   & $ L_{t}^{meD}$                    &$1e-4$     & 2      \\ \hline
2     & Inter   & $ L_{t}^{meRD}$                   &$1e-4$     & 6      \\ \hline
2     & Recon   & $ L_{t}^{recD}$                   &$5e-5$     & 6      \\ \hline
3     & Inter   & $ L_{t}^{meRD}$                   &$1e-4$     & 2      \\ \hline
3     & Recon   & $ L_{t}^{recD}$                   &$5e-5$     & 3      \\ \hline
4     & Recon   & $ L_{t}^{recD}$                   &$5e-5$     & 3      \\ \hline
6     & Recon   & $ L_{t}^{recD}$                   &$5e-5$     & 3      \\ \hline
2     & Recon   & $ L_{t}^{recRD}$                  &$5e-5$     & 6      \\ \hline
3     & Recon   & $ L_{t}^{recRD}$                  &$5e-5$     & 3      \\ \hline
4     & Recon   & $ L_{t}^{recRD}$                  &$5e-5$     & 3      \\ \hline
6     & Recon   & $ L_{t}^{recRD}$                  &$5e-5$     & 3      \\ \hline
2     & All     & $ L_{t}^{all}$                    &$5e-5$     & 15      \\ \hline
3     & All     & $ L_{t}^{all}$                    &$5e-5$     & 15      \\ \hline
4     & All     & $ L_{t}^{all}$                    &$5e-5$     & 15      \\ \hline
6     & All     & $ L_{t}^{all}$                    &$5e-5$     & 10      \\ \hline
6     & All     & $ L_{t}^{all}$                    &$1e-5$     & 10      \\ \hline
6     & All     & $ L_{t}^{all}$                    &$5e-6$     & 5      \\
\bottomrule[1.5pt]
6     & All     & $ L_{T}^{all}$                    &$1e-5$     & 2      \\\hline 
6     & All     & $ L_{T}^{all}$                    &$1e-6$     & 2      \\\hline 
6     & All     & $ L_{T}^{all}$                    &$5e-7$     & 2      \\\hline
6     & All     & $ L_{T}^{all}$                    &$1e-7$     & 4      \\
\bottomrule[1.5pt]
\end{tabular}
\label{table:training_stategy_rgb}
\end{table}
%-------------------------------------------------------------------------
\begin{itemize}
\item As described in (\ref{loss1}), $L_{t}^{meD}$ calculates the distortion $D_{t}^m$ between $x_t$ and its warping frame $\tilde{x}_t$, which is used to obtain the high-fidelity reconstructed motion vectors.
\begin{equation}
    L_{t}^{meD}= w_t \cdot \lambda \cdot D_{t}^m.
\label{loss1}
\end{equation}
\item As described in (\ref{loss2}), based on $L_{t}^{meD}$, $L_{t}^{meRD}$ takes the trade-off between the fidelity $D_{t}^m$ and the consumed bitrate $R_{t}^{m}$ of motion vectors into account. $R_{t}^{m}$ denotes the joint bitrate used for encoding the quantized motion latent representation $\hat{m}_t$ and its associated hyperprior.
\begin{equation}
    L_{t}^{meRD}= w_t \cdot \lambda \cdot D_{t}^{m} + R_{t}^{m}.
\label{loss2}
\end{equation}

\item As described in (\ref{loss3}), $L_{recD}$ calculates the distortion $D_{t}^{y}$ between $x_t$ and its reconstructed frame $\hat{x}_t$, which is used to generate a high-quality reconstructed frame.
\begin{equation}
    L_{t}^{recD}= w_t \cdot \lambda \cdot D_{t}^{y}.
\label{loss3}
\end{equation}

\item As described in (\ref{loss4}),  based on $L_{t}^{recD}$, $L_{t}^{recRD}$ takes the trade-off between the quality of reconstructed frame $\hat{x}_t$ and the consumed bitrate $R_{t}^{y}$ of contextual latent representation $\hat{y}_t$. $R_{t}^{y}$ denotes the joint bitrate used for encoding the quantized contextual latent representation $\hat{y}_t$ and its associated hyperprior.
\begin{equation}
    L_{t}^{recRD}= w_t \cdot \lambda \cdot D_{t}^{y} + R_{t}^{y}.
\label{loss4}
\end{equation}

\item As described in (\ref{loss5}), $L_{t}^{all}$ takes the trade-off between the quality of the reconstructed frame $\hat{x}_t$ and all the consumed bitrate of the coded frame into account.
\begin{equation}
\begin{aligned}
    L_{t}^{all}&= w_t \cdot \lambda \cdot D_{t}^{y} + R_{t}^{m} + R_{t}^{y}.
\end{aligned}
\label{loss5}
\end{equation}

\item As described in (\ref{loss6}), based on $L_{all}$, 
$L_{T}^{all}$ calculates the average loss of multiple frames, which is to achieve a cascaded fine-tuning for reducing the error propagation~\cite{sheng2022temporal,sheng2024spatial,li2022hybrid,li2023neural}.
\begin{equation}
\begin{aligned}
L_{T}^{all}&=\frac{1}{T} \sum_t L_t^{all}\\
&=\frac{1}{T} \sum_t\left\{w_t \cdot \lambda \cdot D_{t}^{y} + R_{t}^{m} + R_{t}^{y}\right\}.
\end{aligned}
\label{loss6}
\end{equation}
\end{itemize}
%-------------------------------------------------------------------------
\par
We use the Lagrangian multiplier $\lambda$ to control the trade-off between the bitrate and distortion. To reduce the error propagation, we follow~\cite{sheng2024spatial, li2023neural} and add a periodically varying weight $w_t$ for each P-frame before the Lagrangian multiplier $\lambda$.
\par

\begin{figure*}[t]
  \centering
  \begin{minipage}[c]{0.3\linewidth}
  \centering
  \includegraphics[width=\linewidth]{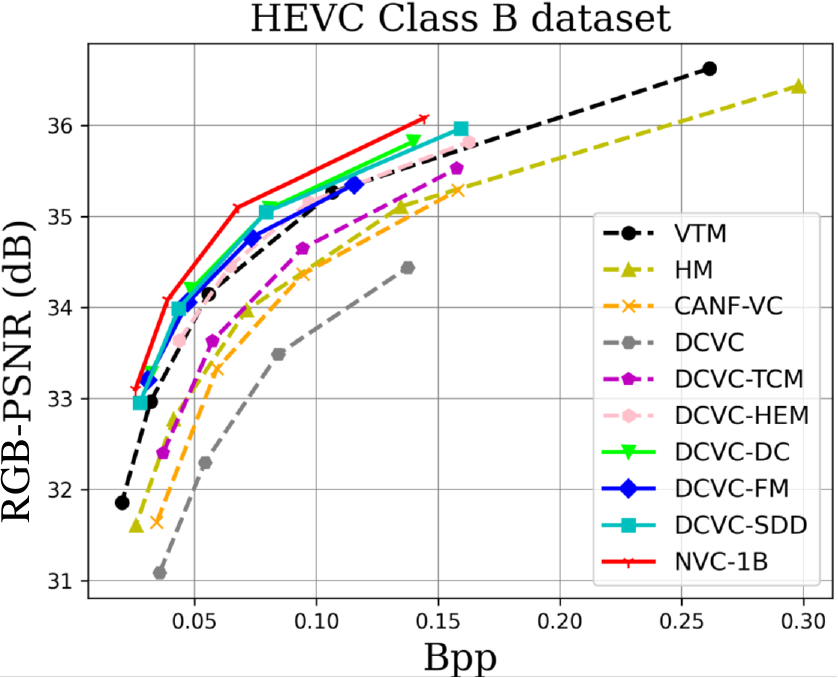}
 \end{minipage}%
  \begin{minipage}[c]{0.3\linewidth}
  \centering
    \includegraphics[width=\linewidth]{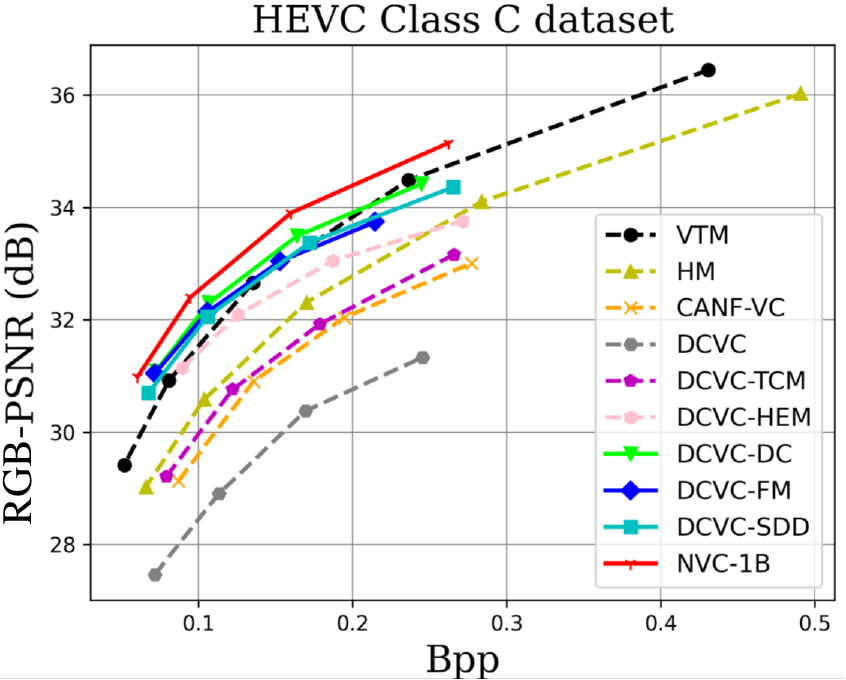}
  \end{minipage}%
  \begin{minipage}[c]{0.3\linewidth}
  \centering
    \includegraphics[width=\linewidth]{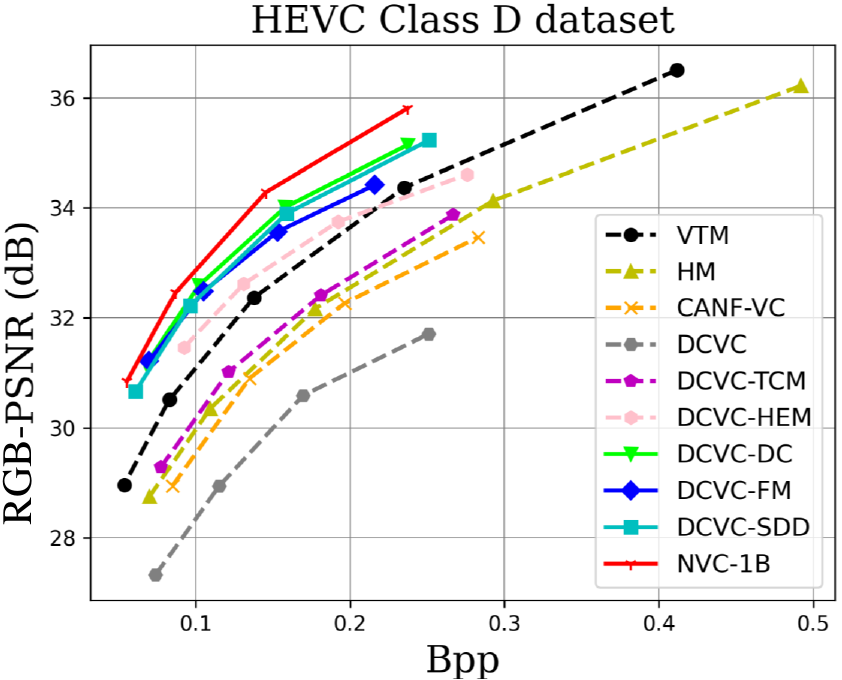}
  \end{minipage}%
  
  \begin{minipage}[c]{0.3\linewidth}
  \centering
    \includegraphics[width=\linewidth]{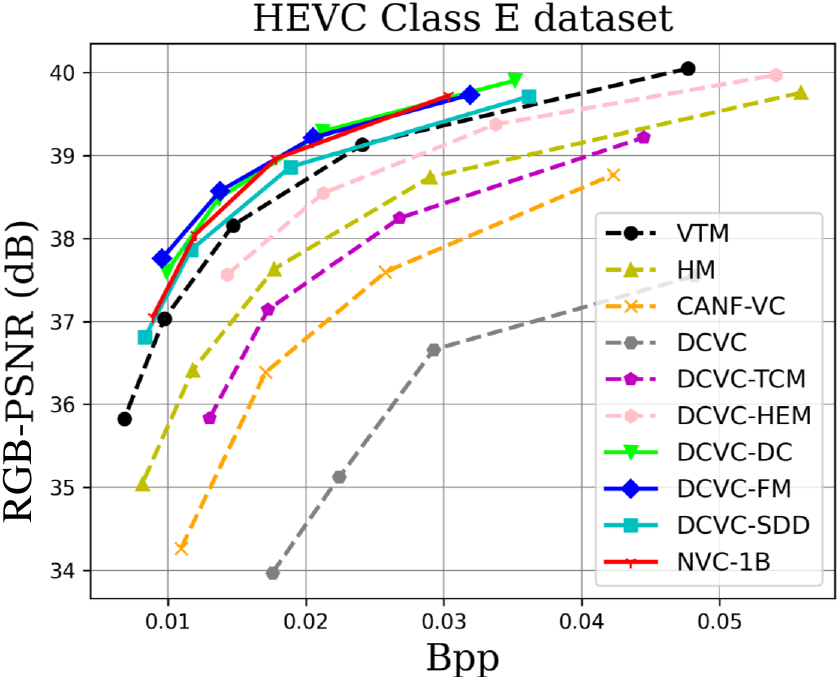}
  \end{minipage}%
  \begin{minipage}[c]{0.3\linewidth}
  \centering
    \includegraphics[width=\linewidth]{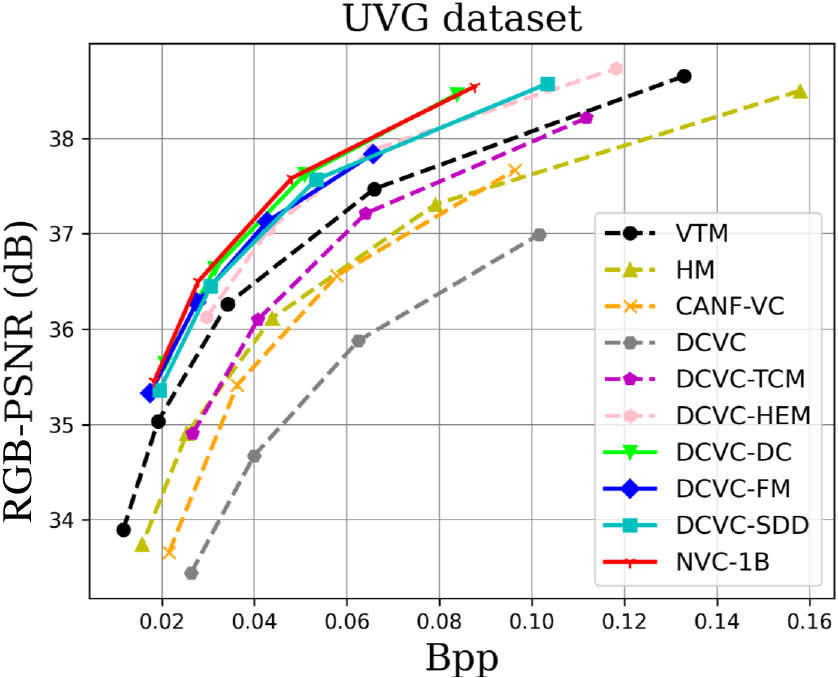}
  \end{minipage}%
  \begin{minipage}[c]{0.3\linewidth}
  \centering
    \includegraphics[width=\linewidth]{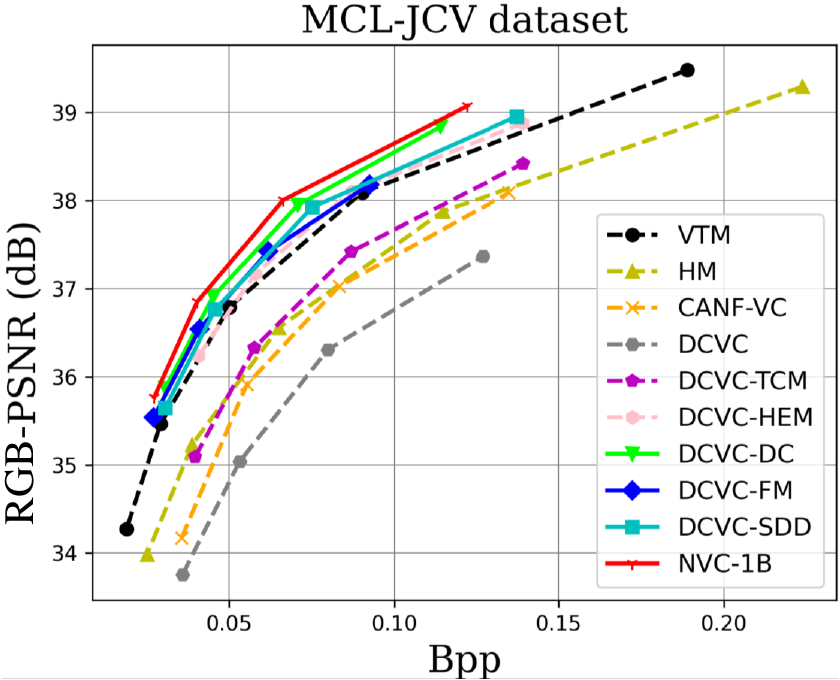}
  \end{minipage}%
    \caption{Rate-distortion curves of the HEVC, UVG, and MCL-JCV RGB video datasets. The reconstruction quality is measured by RGB-PSNR.}
  \label{fig:psnr_results}
\end{figure*}
%------------------------------------------------------------------------
%------------------------------------------------------------------------
\begin{table*}[t]
\caption{BD-rate (\%) comparison in for RGB-PSNR. The anchor is VTM.} 
  \centering
\scalebox{1}{
\begin{threeparttable}
\begin{tabular}{l|c|c|c|c|c|c|c}
\toprule[1.5pt]
               & HEVC Class B  & HEVC Class C  &HEVC Class D &HEVC Class E &UVG             &MCL-JCV  & Average\\ \hline
VTM            &0.0            &0.0            &0.0          &0.0          &0.0             &0.0      &0.0     \\ \hline
HM             &39.0           &37.6           &34.7         &48.6         &36.4            &41.9     &39.7    \\ \hline
CANF-VC        &58.2           &73.0           &48.8         &116.8        &56.3            &60.5     &68.9    \\ \hline
DCVC           &115.7          &150.8          &106.4        &257.5        &129.5           &103.9    &144.0   \\ \hline
DCVC-TCM       &32.8           &62.1           &29.0         &75.8         &23.1            &38.2     &43.5    \\ \hline
DCVC-HEM       &--0.7          &16.1           &--7.1        &20.9         &--17.2          &--1.6    &1.73    \\ \hline
DCVC-DC        &--13.9         &--8.8          &--27.7       &--19.1       &--25.9          &--14.4   &--18.3    \\ \hline
DCVC-FM        &--8.8          &--5.0          &--23.3       &\bf{--20.8}  &--20.5          &--7.4    &--14.3    \\ \hline
DCVC-SDD       &--13.7         &--2.3          &--24.9       &--8.4        &--19.7          &--7.1    &--12.7     \\ \hline
Ours           &\bf{--27.0}    &\bf{--21.2}    &\bf{--37.0}  &--15.4  &\bf{--28.7}     &\bf{--21.3} &\bf{--25.1}   \\ 
\bottomrule[1.5pt]
\end{tabular}
  \begin{tablenotes}
   \item \footnotesize \dag DCVC-SDD~\cite{sheng2024spatial} is our small baseline model.

  \end{tablenotes}
\end{threeparttable}}
\label{table:ip32_psnr}
\end{table*}
%-------------------------------------------------------------------------
%-------------------------------------------------------------------------
\begin{figure*}[t]
  \centering
  \begin{minipage}[c]{0.3\linewidth}
  \centering
  \includegraphics[width=\linewidth]{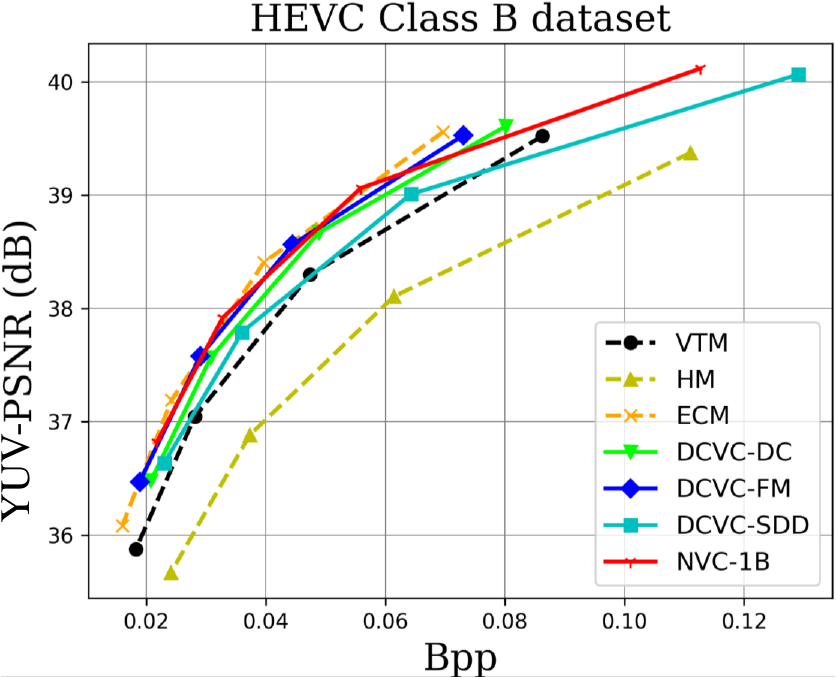}
 \end{minipage}%
  \begin{minipage}[c]{0.3\linewidth}
  \centering
    \includegraphics[width=\linewidth]{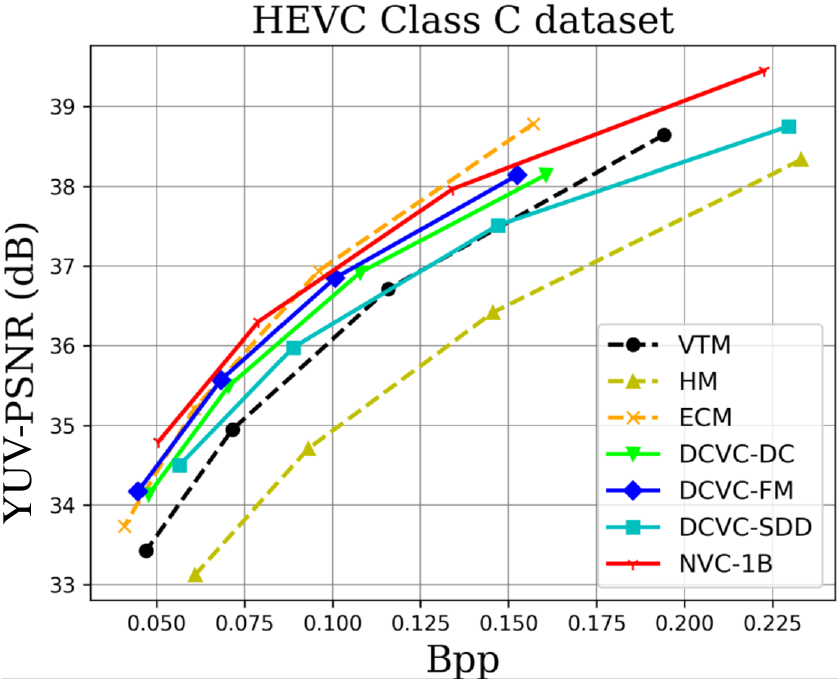}
  \end{minipage}%
  \begin{minipage}[c]{0.3\linewidth}
  \centering
    \includegraphics[width=\linewidth]{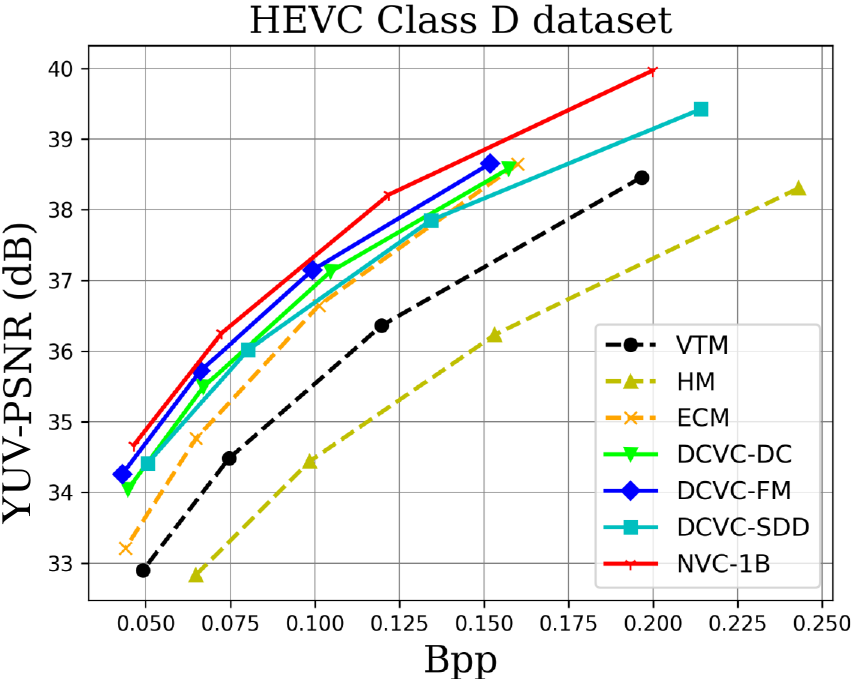}
  \end{minipage}%
  
  \begin{minipage}[c]{0.3\linewidth}
  \centering
    \includegraphics[width=\linewidth]{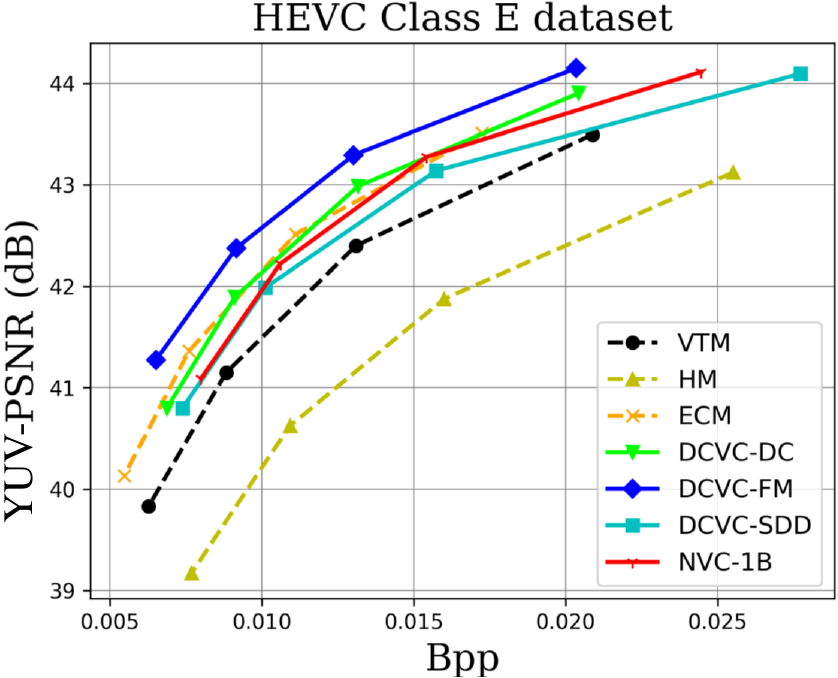}
  \end{minipage}%
  \begin{minipage}[c]{0.3\linewidth}
  \centering
    \includegraphics[width=\linewidth]{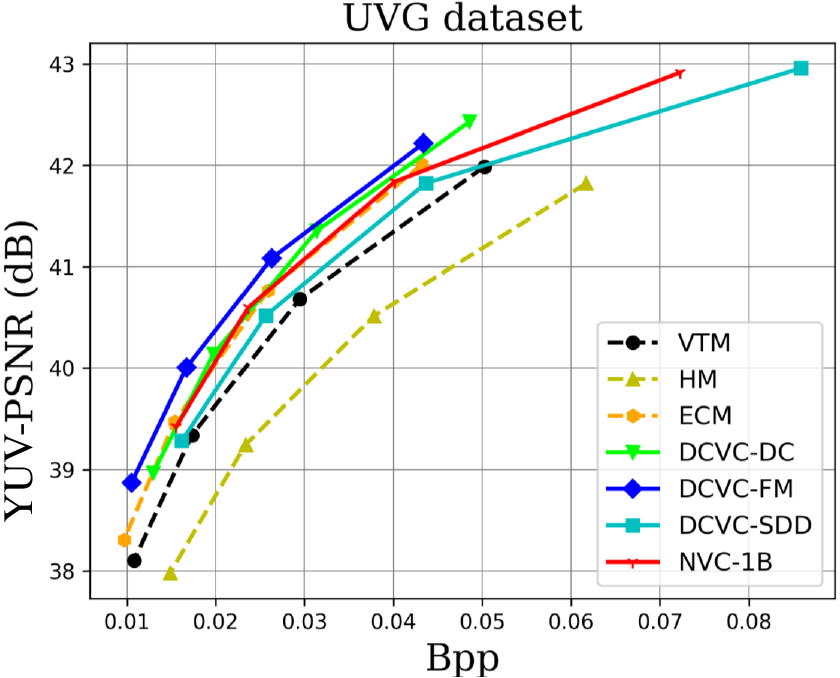}
  \end{minipage}%
  \begin{minipage}[c]{0.3\linewidth}
  \centering
    \includegraphics[width=\linewidth]{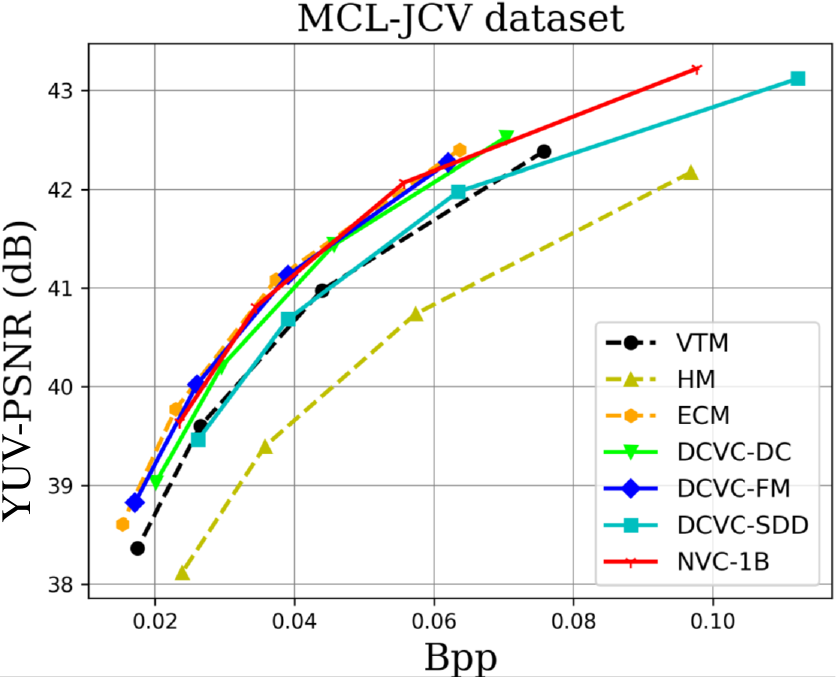}
  \end{minipage}%
    \caption{Rate-distortion curves of the HEVC, UVG, and MCL-JCV YUV420 video datasets. The reconstruction quality is measured by YUV-PSNR.}
  \label{fig:psnr_yuv_results}
\end{figure*}
%------------------------------------------------------------------------
%------------------------------------------------------------------------
\begin{table*}[t]
\caption{BD-rate (\%) comparison in for YUV-PSNR. The anchor is VTM.} 
  \centering
\scalebox{1}{
\begin{threeparttable}
\begin{tabular}{l|c|c|c|c|c|c|c}
\toprule[1.5pt]
               & HEVC Class B  & HEVC Class C  &HEVC Class D &HEVC Class E &UVG             &MCL-JCV  & Average\\ \hline
VTM            &0.0            &0.0            &0.0          &0.0          &0.0             &0.0      &0.0     \\ \hline
ECM            &\bf{--19.5}            &--21.5         &--20.1       &--18.8       &--15.5             &\bf{--18.3}      &--19.0    \\ \hline
HM             &41.1           &36.6           &32.1         &44.7         &38.1            &43.5     &39.4    \\ \hline
DCVC-DC        &--11.6         &--13.1         &--28.8       &--18.1       &--17.2          &--11.0   &--16.6    \\ \hline
DCVC-FM        &--16.6         &--17.7         &--33.4       &\bf{--29.9}        &\bf{--25.0}          &--15.6    &\bf{--23.0}    \\ \hline
DCVC-SDD       &--4.2          &--2.3          &--25.7       &--10.6        &--7.0          &--0.3    &--8.4     \\ \hline
Ours           &--17.7    &\bf{--21.8}    &\bf{--37.4}  &--14.0  &--16.0     &--15.9 &--20.5   \\ 
\bottomrule[1.5pt]
\end{tabular}
  \begin{tablenotes}
   \item \footnotesize \dag DCVC-SDD~\cite{sheng2024spatial} is our small baseline model.
   \item \footnotesize \dag\dag Since ECM-13.0~\cite{ECM} only supports encoding YUV420 videos currently, we only present its compression results in this table.
  \end{tablenotes}
\end{threeparttable}}
\label{table:ip32_yuv_psnr}
\end{table*}
%-------------------------------------------------------------------------

%-------------------------------------------------------------------------
%-------------------------------------------------------------------------
\begin{figure*}[t]
  \centering
   \includegraphics[width=0.85\linewidth]{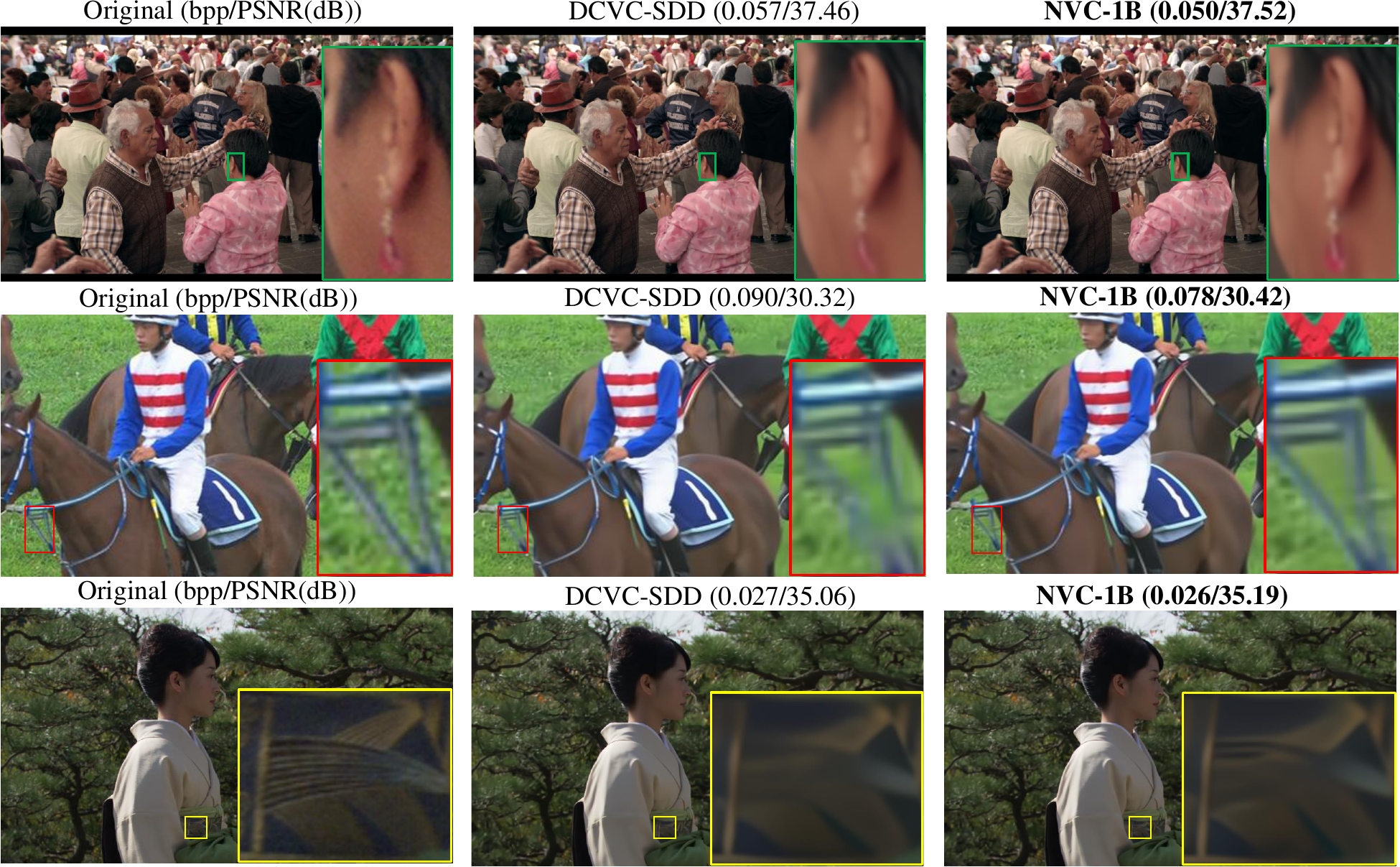}
      \caption{Subjective quality comparison between our proposed large model---NVC-1B and its small baseline model---DCVC-SDD~\cite{sheng2024spatial} on the 5th frame of the MCL-JCV {\em videoSRC14} sequence, the 7th frame of the HEVC Class D {\em RaceHorses} sequence and the 17th frame of the HEVC Class B {\em Kimono1} sequence.}
   \label{fig:subjective}
\end{figure*}
%-------------------------------------------------------------------------
%-------------------------------------------------------------------------
\begin{table}[t]
\caption{average encoding/decoding time for a 1080P frame (in seconds).} 
  \centering
\scalebox{1}{
\begin{threeparttable}
\begin{tabular}{c|c|c}
\toprule[1.5pt]
Schemes                            & Enc Time   & Dec Time                 \\ \hline
VTM                                & 743.88 s   & 0.31 s                   \\ \hline
ECM                                & 5793.88 s        & 1.28 s                      \\ \hline
HM                                 & 92.58 s    & 0.21 s                   \\ \hline
DCVC-SDD   & 0.89 s     & 0.70 s                   \\ \hline
Ours                               & 4.44 s     & 3.54 s                   \\ 
\bottomrule[1.5pt]
\end{tabular}

  %\vspace{-0.5cm}
\end{threeparttable}}
\label{table:enc_dec_time}
\end{table}

%-------------------------------------------------------------------------

\section{Experiments}\label{sec:experiments}
\subsection{Experimental Setup}
\subsubsection{Training and Testing Data}
For training, we use 7-frame videos of the Vimeo-90k~\cite{xue2019video} dataset.  Following most existing neural video coding models~\cite{li2021deep,sheng2022temporal,li2022hybrid,sheng2024spatial}, we randomly crop the original videos into 256$\times$256 patches for data augmentation. 
For testing, we use HEVC dataset~\cite{bossen2013common}, UVG dataset~\cite{mercat2020uvg}, and MCL-JCV dataset~\cite{wang2016mcl} in RGB and YUV420 format. These datasets contain videos with different contents, motion patterns, and resolutions, which are commonly used to evaluate the performance of neural video coding models. When testing RGB videos, we convert the videos in YUV420
format to RGB format using FFmpeg. 

\subsubsection{Implementation Details}
Following~\cite{sheng2024spatial, li2023neural}, we set 4 base $\lambda$ values (85, 170, 380, 840) to control the rate-distortion trade-off. For hierarchical quality, we set the periodically varying weight $w_t$ before $\lambda$ as (0.5, 1.2, 0.5, 0.9). We implement our NVC-1B model with PyTorch. AdamW~\cite{kingma2014adam} is used as the optimizer and batch size is set to 32. Before the multi-frame cascaded fine-tuning stage, we train the NVC-1B model on 32 NVIDIA Ampere Tesla A40 (48G memory) GPUs for 85 days. In the multi-frame cascaded fine-tuning stage, we train the NVC-1B model on 4 NVIDIA A800 PCIe (80G memory) GPUs for 28 days. To alleviate the CUDA memory pressure induced by multi-frame cascaded training, Forward Recomputation Backpropagation (FRB)\footnote{\url{https://qywu.github.io/2019/05/22/explore-gradient-checkpointing.html}} is used.

\subsubsection{Test Configurations}
As with most previous neural video coding models, we focus on the low-delay coding scenario in this paper. Following~\cite{sheng2022temporal, sheng2024spatial,li2023neural,li2021deep}, we test 96 frames for each video sequence and set the intra-period to 32.
For traditional video codecs, we choose HM-16.20~\cite{HM}, VTM-13.2~\cite{VTM}, and ECM-5.0~\cite{ECM} as our benchmarks. HM-16.20 is the official reference software of H.265/HEVC.  VTM-13.2 is the official reference software of H.266/VVC. ECM is the prototype of the next-generation traditional codec. We use \emph{encoder\_lowdelay\_main (\_rext)}, \emph{encoder\_lowdelay\_vtm}, , and \emph{encoder\_lowdelay\_ecm} configurations for HM-16.20, VTM-13.2, and ECM-13.0, respectively. The detailed commands for HM-16.20, VTM-13.2, and ECM-13.0 are shown as follows.
\begin{itemize}
    \item -c $\{\emph{config file name}\}$  \mbox{-}\mbox{-}InputFile=$\{\emph{input file name}\}$ \mbox{-}\mbox{-}InputChromaFormat=$\{\emph{input chroma format}\}$ \mbox{-}\mbox{-}FrameRate=$\{\emph{frame rate}\}$ \mbox{-}\mbox{-}DecodingRefreshType=2 \mbox{-}\mbox{-}InputBitDepth=8 \mbox{-}\mbox{-}FramesToBeEncoded=96 \mbox{-}\mbox{-}SourceWidth=$\{\emph{width}\}$  \mbox{-}\mbox{-}SourceHeight=$\{\emph{height}\}$ 
    \mbox{-}\mbox{-}IntraPeriod=32 \mbox{-}\mbox{-}QP=$\{\emph{qp}\}$ \mbox{-}\mbox{-}Level=6.2 \mbox{-}\mbox{-}BitstreamFile=$\{\emph{bitstream file name}\}$
\end{itemize}
For neural video coding models, we choose CANF-VC~\cite{ho2022canf}, DCVC~\cite{li2021deep}, DCVC-TCM~\cite{sheng2022temporal}, DCVC-HEM~\cite{li2022hybrid}, DCVC-DC~\cite{li2022hybrid}, DCVC-FM~\cite{li2024neural}, and our small baseline model---DCVC-SDD~\cite{sheng2024spatial} as our benchmarks. 

\subsubsection{Evaluation Metrics}
When testing RGB videos, we use RGB-PSNR as the distortion evaluation metric. When testing YUV420 videos, following DCVC-DC~\cite{li2022hybrid}, we use the compound YUV PSNR as the distortion evaluation metric. The weight of YUV components is set to 6:1:1~\cite{ohm2012comparison}. 
 Bits per pixel (bpp) is used as the bitrate evaluation metric. BD-rate~\cite{bjontegaard2001calculation} is used to compare the compression performance of difference models, where negative numbers indicate bitrate saving and positive numbers indicate bitrate increasing.\par
%-------------------------------------------------------------------------
\begin{figure*}[t]
  \centering
   \includegraphics[width=0.9\linewidth]{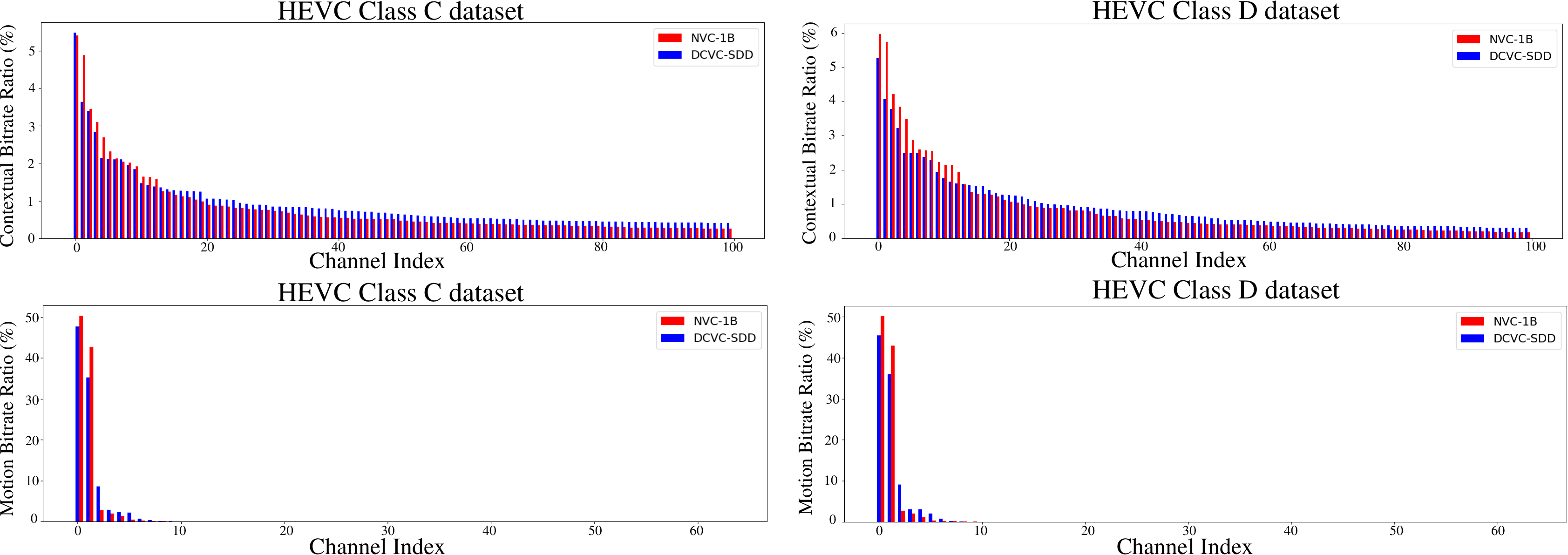}
      \caption{Comparison of the average contextual and motion channel bitrate distribution between our proposed large model---NVC-1B and its small baseline model---DCVC-SDD on the HEVC Class C and D datasets.}
   \label{fig:energy_compaction}
\end{figure*}
%------------------------------------------------------------------------
\subsection{Experimental Results}
\subsubsection{Objective Comparison Results for RGB Videos}
When testing RGB videos, we illustrate the rate-distortion curves on the HEVC, UVG, and MCL-JCV RGB video datasets in Fig.~\ref{fig:psnr_results}. The curves show that our proposed large neural video coding model---NVC-1B has significantly outperformed its small baseline model---DCVC-SDD. We list the detailed BD-rate comparison results in Table.~\ref{table:ip32_psnr}. The results show that our NVC-1B achieves an average --25.1\% BD-rate reduction over VTM-13.2 across all test datasets, which is much better than that of DCVC-SDD (--12.7\%). It even surpasses the state-of-the-art neural video coding models---DCVC-DC (--18.3\%) and DCVC-FM (--14.3\%). \par
When testing YUV420 videos, we illustrate the rate-distortion curves on the YUV420 videos in Fig.~\ref{fig:psnr_yuv_results} and list the corresponding BD-rate comparison results in Table.~\ref{table:ip32_yuv_psnr}. The results show that based on our small baseline model---DCVC-SDD, the large model improves the average compression performance from --8.4\% to --20.5\%, which even outperforms ECM (--19.0\%). \par
However, we find that the performance gain on the HEVC Class E dataset is smaller than that of other datasets. This is mainly because the number of frames used for multi-frame cascaded fine-tuning is small (6 frames) under the limitation of GPU CUDA memory. If we have GPUs with larger CUDA memory, we can use more frames (DCVC-DC uses 7 frames and DCVC-FM uses 32 frames) to fine-tune our large model to improve its compression performance.

\subsubsection{Subjective Comparison Results}
We visualize the reconstructed frames of our proposed NVC-1B and its small baseline model---DCVC-SDD in Fig.~\ref{fig:subjective}. By comparing their subjective qualities, we observe that the frames reconstructed by NVC-1B can retain more textures with a lower bitrate. We take the {\em videoSRC14} sequence of the MCL-JCV dataset, the {\em RaceHorses} sequence of the HEVC Class D dataset, and the {\em Kimono1} sequence of the HEVC Class B dataset as examples. For the {\em videoSRC14} sequence, our NVC-1B model can use the bitrate of 0.051 bpp (0.057 bpp for DCVC-SDD) to achieve a reconstructed frame of 37.52 dB (37.46 dB for DCVC-SDD). Observing the earring of the dancing woman in {\em videoSRC14}, we can find our NVC-1B model can keep sharper edges. For the {\em RaceHorses} sequence, our NVC-1B model can use the bitrate of 0.078 bpp (0.090 bpp for DCVC-SDD) to achieve a reconstructed frame of 30.42 dB (30.32 dB for DCVC-SDD). Comparing the horse's reins, we can see that our NVC-1B model can reconstruct the reins but DCVC-SDD cannot. For the {\em Kimino1} sequence,  our NVC-1B model can use the bitrate of 0.026 bpp (0.027 bpp for DCVC-SDD) to achieve a reconstructed frame of 35.19 dB (35.06 dB for DCVC-SDD). Zooming in the belt on the woman, the frame reconstructed by our NVC-1B model can retain more details.
%-------------------------------------------------------------------------
%-------------------------------------------------------------------------
\begin{figure}[t]
  \centering
   \includegraphics[width=0.95\linewidth]{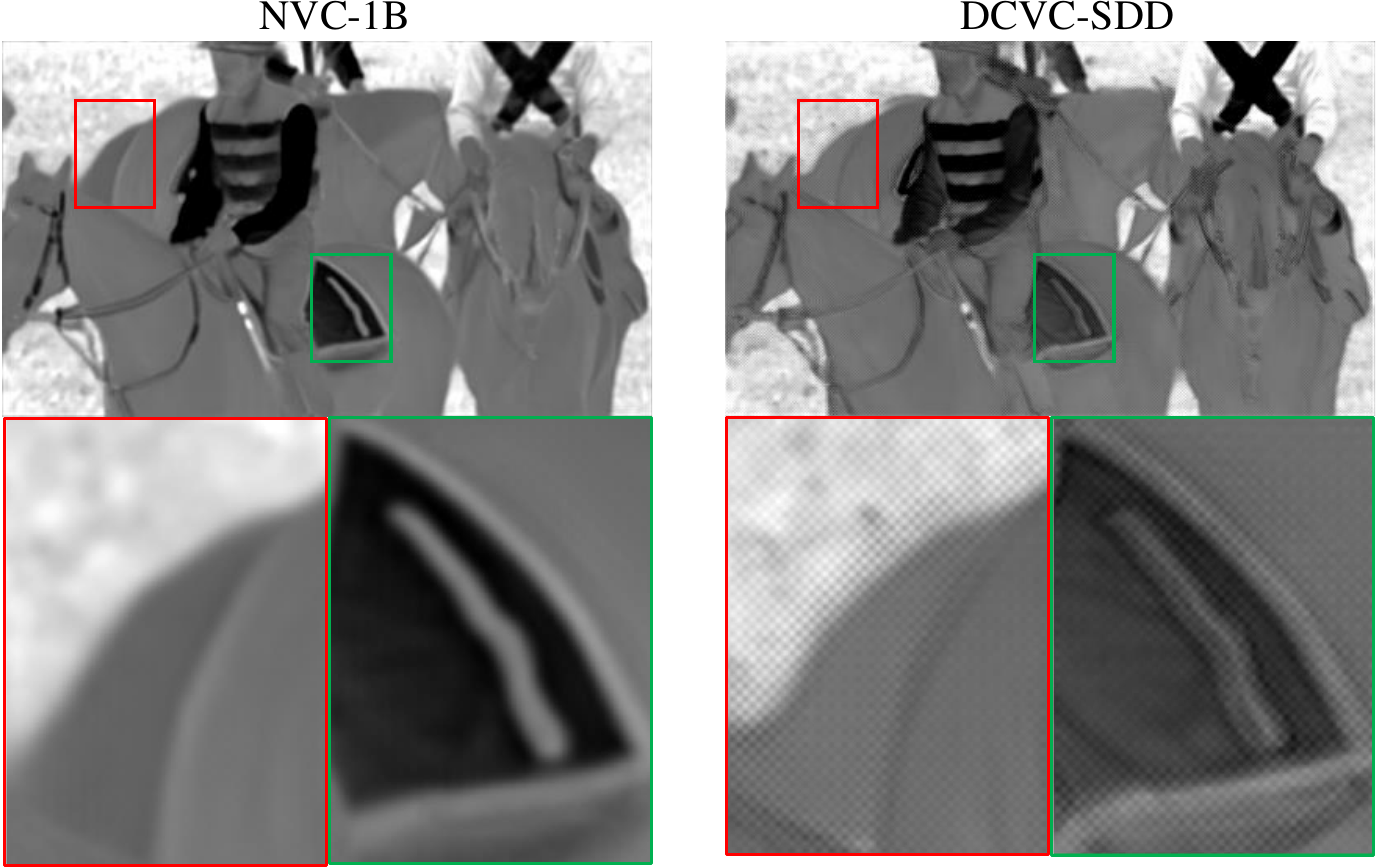}
      \caption{Comparison of the temporal contexts predicted by our proposed large model---NVC-1B and its small baseline model---DCVC-SDD.}
   \label{fig:TCM_visualization}
\end{figure}
%------------------------------------------------------------------------
\subsubsection{Encoding/Decoding Time} Following the setting of our small baseline model---DCVC-SDD, we include the time for model inference, entropy modeling, entropy coding, and data transfer between CPU and GPU. All the learned video codecs are run on a NVIDIA A800 GPU. As shown in Table~\ref{table:enc_dec_time}, the encoding and decoding time of our NVC-1B model for 1080p video are 4.44s and 3.54s per frame, respectively, which are five times longer than that of DCVC-SDD. However, the encoding time of our NVC-1B model is significantly lower than that of traditional video codecs. With the development of lightweight techniques~\cite{hu2021lora,lu2023flexgen} for large models, the encoding and decoding time can be further optimized.

\subsection{Analysis}
\subsubsection{Analysis of Transform Energy Compaction}\label{ablation1}
To explore why our proposed large video coding model---NVC-1B can bring performance gain, we analyze the transform energy compaction. We calculate the average bitrate ratio of each channel of the contextual latent representation $\hat{y}_t$ over the HEVC Class C and D datasets. Specifically, the contextual bitrate ratio (CR) of channel $c$ for each video sequence is first calculated as (\ref{loss7}).
\begin{equation}
\begin{aligned}
CR&=\frac{\sum_t \left\{-\log_2(p(\hat{y}_t^c))\right\}}{\sum_c \sum_t 
 \left\{-\log_2(p(\hat{y}_t^c))\right\}}.
\end{aligned}
\label{loss7}
\end{equation}
Then, we calculate the average contextual bitrate ratio over all sequences. 
We sort the average contextual bitrate ratio of each channel from largest to smallest to analyze the transform energy compaction. As illustrated in Fig.~\ref{fig:energy_compaction}, for contextual bitrate ratio, we present the top 100 channels with obvious bitrate overhead. Comparing the channel bitrate distribution, we find that the contextual latent representations generated by our proposed large video coding model have higher transform energy compaction than our small baseline model---DCVC-SDD. More bitrates are concentrated in fewer channels. \par
In addition to the contextual latent representations, we also calculate the channel bitrate ratio of motion latent representations $\hat{m}_t$.  Similarly, the motion bitrate ratio (MR) of channel $c$ for each video sequence is first calculated as (\ref{loss8}).
\begin{equation}
\begin{aligned}
MR &=\frac{\sum_t \left\{-\log_2(p(\hat{m}_t^c))\right\}}{\sum_c \sum_t 
 \left\{-\log_2(p(\hat{m}_t^c))\right\}}.
\end{aligned}
\label{loss8}
\end{equation}
Then, we calculate the average motion bitrate ratio over all sequences. As presented in Fig.~\ref{fig:energy_compaction}, by ranking the average motion bitrate ratio of each channel, we find that although we allocate most of the parameters to contextual parts, the transform energy compaction can be improved for motion compression by end-to-end training. The motion channel bitrate distribution of our large model is also more concentrated. 
% The analysis indicates that our large neural video coding can obtain higher compression performance by increasing the transform energy compaction.
The analysis indicates that our large video coding model can obtain a higher compression performance by increasing the transform energy compaction.
% More signal energy can be concentrated in fewer channels, which is beneficial for entropy coding.
More signal energy is concentrated in fewer channels, which is beneficial for entropy coding.

\subsubsection{Analysis of Temporal Context Mining Accuracy}\label{ablation2}
We further analyze the temporal context mining accuracy to explore why our proposed large video coding model---NVC-1B can bring performance gain. We visualize the temporal context $C_t^0$ predicted by NVC-1B and its small baseline model---DCVC-SDD in Fig.~\ref{fig:TCM_visualization}. By zooming in the temporal contexts, we find that the context predicted by NVC-1B has smoother object edges, whereas that predicted by DCVC-SDD has obvious prediction artifacts. For example, the edges of the horse's tail and saddle predicted by DCVC-SDD have much noise, which seriously reduces the temporal prediction accuracy. The analysis indicates that our NVC-1B model can obtain a higher compression performance by improving the temporal context mining accuracy. 
%-------------------------------------------------------------------------
\begin{figure}[t]
  \centering
   \includegraphics[width=0.75\linewidth]{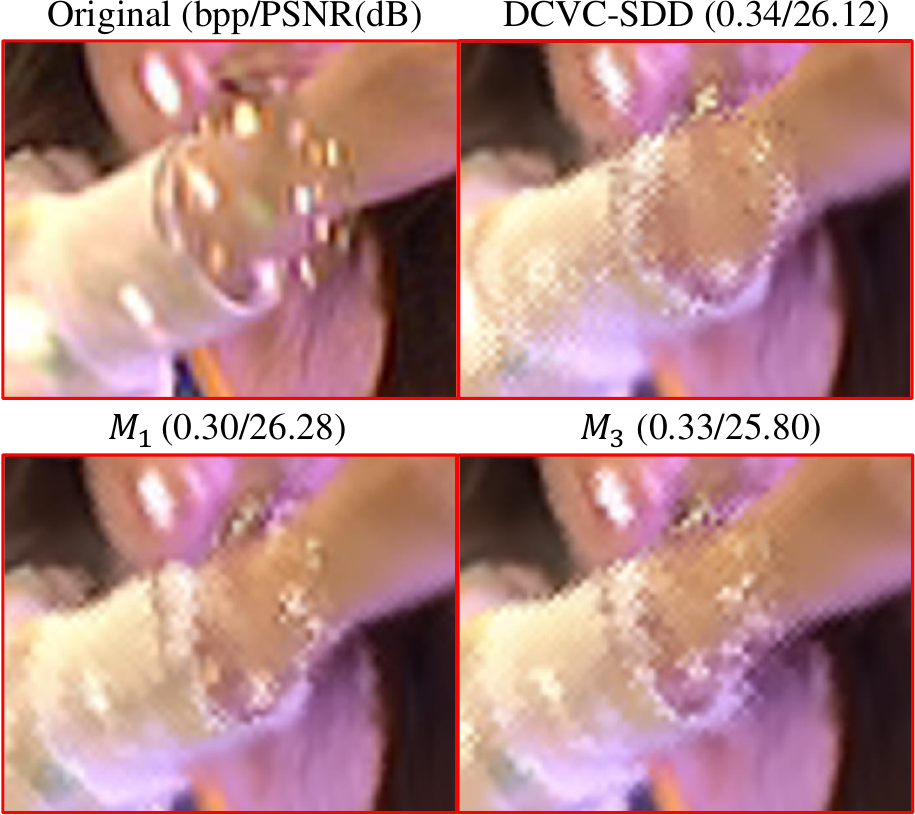}
      \caption{Comparison of the warp frames predicted by the small baseline model---DCVC-SDD, $M_1$, and $M_3$ models mentioned in Section~\ref{sec:mv_encoder_decoder}.}
   \label{fig:mv_warpframe}
\end{figure}
%------------------------------------------------------------------------
\subsubsection{Analysis of Scaling Up for Motion Encoder-Decoder}\label{ablation3}
In Section~\ref{sec:mv_encoder_decoder}, we observe a strange phenomenon that if we slightly increase the model size of the motion encoder and decoder it can lead to performance gains. However, the performance gain will decrease if we continuously increase their model size. To explore the reason for this strange phenomenon, we visualize the warp frames predicted by the small baseline model---DCVC-SDD, $M_1$, and $M_3$ models mentioned in Section~\ref{sec:mv_encoder_decoder}. As illustrated in Fig.~\ref{fig:mv_warpframe}, we find that, at a similar bitrate,  the $M_1$ model with 52M parameters can obtain a more accurate warp frame than DCVC-SDD with 21M parameters. The arm and bubbles predicted by the $M_1$ model have fewer artifacts. However, the accuracy of the warp frame predicted by $M_3$ decreases dramatically. Its prediction artifacts of arms and bubbles increase significantly. The analysis indicates that the model size of the motion encoder-decoder is not always the bigger the better. An inappropriate model size will reduce the accuracy of temporal prediction.

\section{Conclusion}\label{sec:conclusion}
In this paper, we scale up the model sizes of different parts of a neural video coding model to analyze the influence of model size on compression performance. In addition, we use different architectures to implement a neural video coding model to analyze the influence of model architecture architecture on compression performance. Based on our exploration, we design the first large neural video coding model---NVC-1B. Experimental results show that our proposed NVC-1B model can achieve a significant video compression gain over its small baseline model. In the future, we will design larger models with more efficient architectures to further improve the video compression performance.

\bibliographystyle{ieeetr}
\bibliography{ref}
\end{document}